\documentclass[sigconf]{acmart}

\usepackage{latexsym}
\usepackage{enumitem}
\usepackage{amsthm}
\usepackage{amsmath}
\usepackage{booktabs}
\usepackage{float}
\usepackage{url}
\usepackage{graphicx,verbatimbox}
\usepackage{subcaption}
\usepackage{stfloats}
\usepackage[ruled,noend]{algorithm2e}
\SetAlFnt{\small}

\newcommand{\prob}{\mathbb{P}}

\newcommand{\bh}{\mathbf{h}}
\newcommand{\bc}{\mathbf{c}}
\newcommand{\bv}{\mathbf{v}}

\newtheorem{thm:eg}{Example}

\newcommand{\etc}{\emph{etc.}\xspace} 
\newcommand{\ie}{\emph{i.e.}\xspace} 
\newcommand{\eg}{\emph{e.g.}\xspace} 
\newcommand{\eq}{\emph{Eq.}\xspace} 
\newcommand{\nop}[1]{}

\newcommand{\wx}[1]{{#1}}

\newcommand{\NERO}{\textsc{Nero}\xspace}

\SetKwInput{KwInput}{Input}
\SetKwInput{KwOutput}{Output}
\SetEndCharOfAlgoLine{}

\AtBeginDocument{%
  \providecommand\BibTeX{{%
    \normalfont B\kern-0.5em{\scshape i\kern-0.25em b}\kern-0.8em\TeX}}}

\copyrightyear{2020}
\acmYear{2020}
\setcopyright{rightsretained}

\acmConference{The Web Conference '20}{April 20 - 24, 2020}{Taipei}
\acmDOI{10.1145/330ccc6307.3328180}
\acmISBN{978-1-4503-6317-4/19/07}
\acmBooktitle{WWW '20, April 20 - 24, 2020, Taipei}
\usepackage[normalem]{ulem}
\usepackage{amsmath}
\useunder{\uline}{\ul}{}

\begin{document}
\title{NERO: A Neural Rule Grounding Framework for\\ Label-Efficient Relation Extraction}

\author{Wenxuan Zhou$^1$, Hongtao Lin$^1$, Bill Yuchen Lin$^1$, Ziqi Wang$^2$}
\author{Junyi Du$^1$, Leonardo Neves$^3$, Xiang Ren$^1$}
\affiliation{\textsuperscript{1}University of Southern California  \textsuperscript{2}Tsinghua University  \textsuperscript{3}Snapchat Inc. \\
\textsuperscript{1}\{zhouwenx, lin498, yuchen.lin, junyidu, xiangren\}@usc.edu~~ \textsuperscript{2}ziqi-wan16@mails.tsinghua.edu.cn~~\textsuperscript{3}lneves@snap.com
}

\renewcommand{\authors}{Wenxuan Zhou, Hongtao Lin, Bill Yuchen Lin, Ziqi Wang, Junyi Du, Leonardo Neves, and Xiang Ren}
\renewcommand{\shortauthors}{W. Zhou et al.}

\begin{abstract}
Deep neural models for relation extraction tend to be less reliable when perfectly labeled data is limited, despite their success in label-sufficient scenarios.
Instead of seeking more instance-level labels from human annotators, here we propose to annotate frequent surface patterns to form \textit{labeling rules}. These rules can be automatically mined from large text corpora and generalized via a soft rule matching mechanism.
Prior works use labeling rules in an exact matching fashion, which inherently limits the coverage of sentence matching and results in the low-recall issue. 
In this paper, we present a neural approach to ground rules for RE, named \NERO, which jointly learns a relation extraction module and a soft matching module. 
One can employ any neural relation extraction models as the instantiation for the RE module.
The soft matching module learns to match rules with \textit{semantically similar} sentences such that raw corpora can be automatically labeled and leveraged by the RE module (in a much better coverage) as augmented supervision, in addition to the exactly matched sentences.
Extensive experiments and analysis on two public and widely-used datasets demonstrate the effectiveness of the proposed \NERO framework, comparing with both rule-based and semi-supervised methods.
Through user studies, we find that the time efficiency for a human to annotate rules and sentences are similar (0.30 vs. 0.35 min per label). In particular, \NERO's performance using 270 rules is comparable to the models trained using 3,000 labeled sentences, yielding a 9.5x speedup.
Moreover, \NERO can predict for unseen relations at test time and provide interpretable predictions.
We release our code\footnote{\url{https://github.com/INK-USC/NERO}} to the community for future research.
\end{abstract}

\maketitle

\begin{figure}
    \centering
    \includegraphics[width=0.9\linewidth]{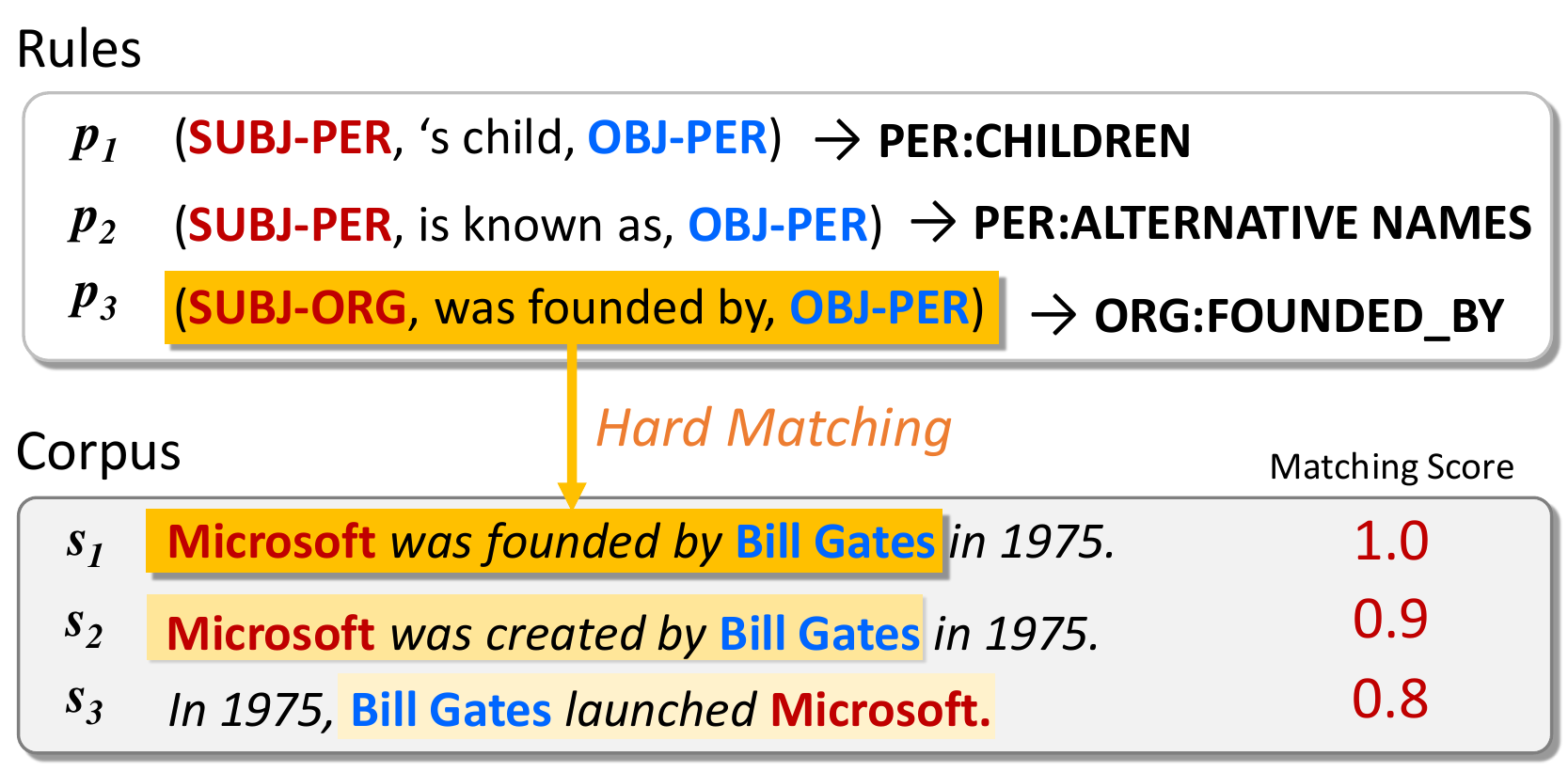}
    \vspace{-0.2cm}
    \caption{
    Current rule-based methods mostly rely on exact/hard matching to raw corpus and suffer from limited coverage. For example, the rule body of $p_3$ only matches sentence $s_1$ but is also similar to $s_2$ and $s_3$, which express the same relation as $p_3$. A ``soft rule matching" mechanism is desirable to make better use of the corpus for label generation.
    }
    \label{fig:rule_eg}
    \vspace{-0.3cm}
\end{figure}

\section{Introduction}
\label{sec:intro}
Relation extraction (RE) plays a key role in information extraction tasks and knowledge base construction, which aims to identify the relation between two entities in a given sentence.
For example, given the sentence ``Bill Gates is the founder of Microsoft'' and an entity pair (``Bill Gates'', ``Microsoft''), a relation classifier is supposed to predict the relation of \texttt{ORG:FOUNDED\_BY}. 
Recent advance in neural language processing has shown that neural models \cite{zeng2015distant, zhang2017position, zhang2018graph} gained great success on this task, yielding state-of-the-art performance when a large amount of well-annotated sentences are available.
However, these supervised learning methods degrade dramatically when the sentence-level labels are insufficient. 
This problem is partially solved by the use of distant supervision based on knowledge bases (KBs) \cite{mintz2009distant, surdeanu2012multi}.
They usually utilize an existing KB for automatically annotating an unlabeled corpus with an over-simplified assumption --- two entities co-occurring in a sentence should be labeled as their relations in the KB regardless of their contexts. 
While distant supervision automates the labeling process, it also introduces noisy labels due to context-agnostic labeling, which can hardly be eliminated.

In addition to KBs, labeling rules 
are also important means of representing domain knowledge~\cite{ratner2016data}, 
which can be automatically mined from large corpora~\cite{nakashole2012patty, jiang2017metapad}. 
Labeling rules can be seen as a set of pattern-based heuristics for matching a sentence to a relation. These rules are much more accurate than using KBs as distant supervision. 
\wx{The traditional way of using such rules is to perform exact string matching, and a sentence is either able or unable to be matched by a rule.
This kind of hard-matching method inherently limits the generalization of the rules for sentences with similar semantics but dissimilar words, which consequently causes the low-recall problem and data-insufficiency for training neural models.
For example, in Fig. \ref{fig:rule_eg}, the rule $P_3$ can only find $S_1$ as one hard-matched instance for training the model while other sentences (e.g., $S_2$ and $S_3$) should also be matched for they have a similar meaning.} 

\begin{figure}
\vspace{-0.2cm}
\hspace{0.2cm}
    \centering
    \includegraphics[width=0.95\linewidth]{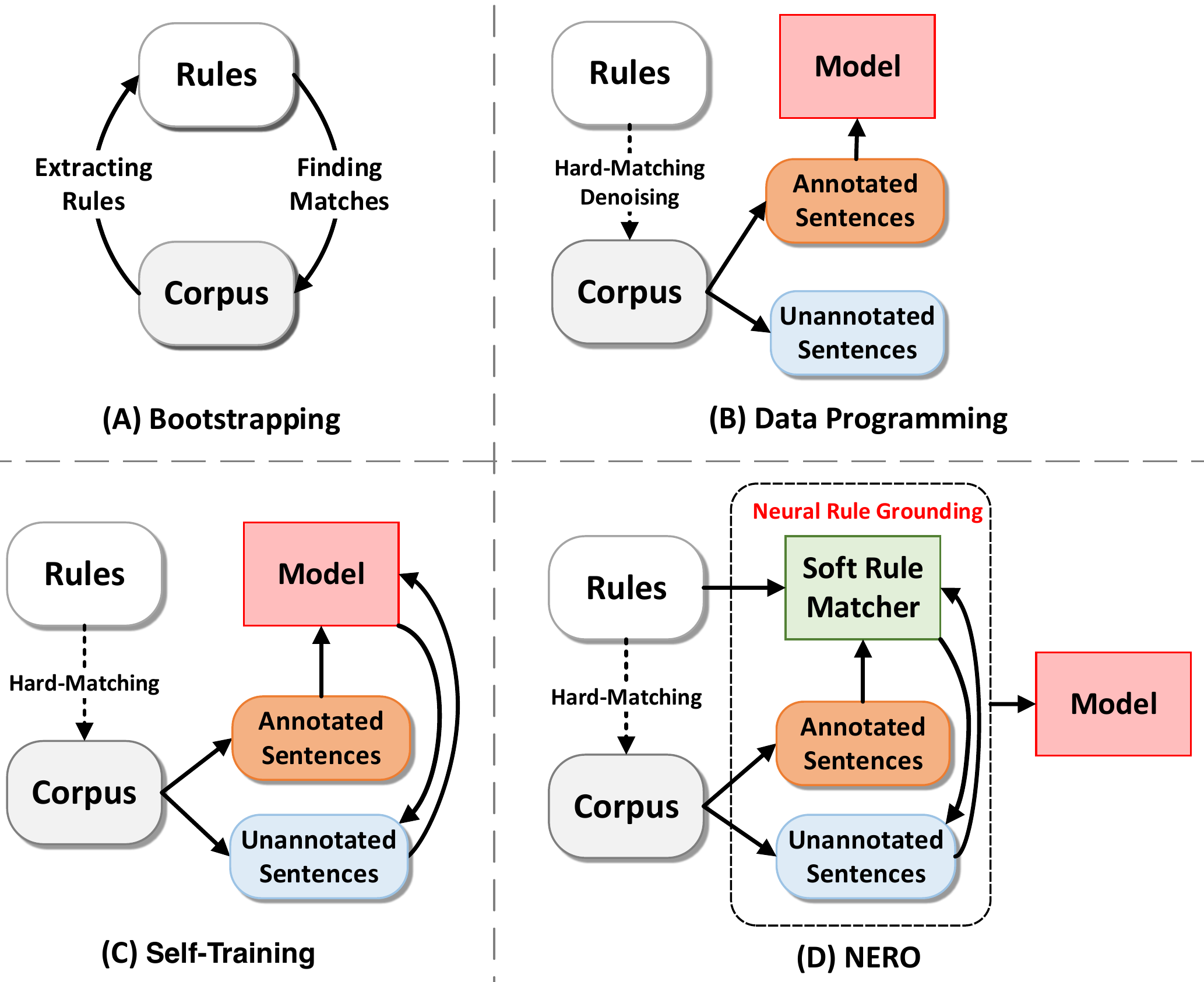}
    \vspace{-0.3cm}
    \caption{\textbf{Comparison between previous work and the proposed \NERO framework.} (A) Bootstrapping. (B) Data Programming. (C) Self-Training. (D) \NERO. The neural rule grounding process enables the soft-matching between rules and unmatched sentences using the soft rule matcher.}
    \vspace{-0.3cm}
    \label{fig::framework}
\end{figure}
Many attempts have been made to RE using labeling rules and unlabeled corpora, as summarized in Fig.~\ref{fig::framework}. 
\textit{Rule-based bootstrapping} methods \cite{jones1999bootstrapping, agichtein2000snowball, batista2015semi} extract relation instances from the raw corpus by a pre-defined rule mining function (e.g., TF-IDF, CBOW) and expanding the rule set in an iterative way to increase the coverage. 
However, they still make predictions by performing hard-matching
on rules, which is context-agnostic and suffers from the low-recall problem. 
Also, their matching function is not learnable and thus has trouble in extracting semantically similar sentences.
Another typical option is \textit{data programming} \cite{ratner2016data, Hancock_2018}, which aims to annotate the corpus using rules with model fitting.
It trains a neural RE model using the hard-matched sentences by rules and then reduces the noise of rules with their proposed algorithms. 
However, it does not consider the massive data that fail to be annotated by hard-matching.
\textit{Self-training} \cite{rosenberg2005semi}, as a semi-supervised framework, does attempt to utilize unmatched sentences by using confident predictions of a learned model.
Then, they train the model again and again by iteratively generating more confident predictions. 
However, it does not explicitly model the soft matching ability of rules over unmatched sentences, making the generated labels noisy and unreliable.

In this paper, we want to explicitly exploit labeling rules over unmatched sentences as supervision for training better RE models, for which we propose a NEural Rule grOunding (\NERO) framework.
The \NERO framework has two major modules: a sentence-level relation classifier and a soft rule matcher.
The former aims to learn the neural representations of sentences and classify which relation it talks about, which serves as the outcome of \NERO.
We first apply our collected rules on a raw corpus by \textit{hard matching} and use hard-matched sentences as the main training data for \NERO.
The unmatched sentences are assigned with pseudo labels by the soft rule matcher, which is a \textit{learnable} module that produces matching scores for unmatched sentences with collected rules.
The key intuition behind our soft rule matcher is that the distances between rules and sentences can be modeled by simple cosine computations in a new space transformed from their neural representations (e.g. word embeddings),
which can be learned by a contrastive loss.
Jointly training the two modules reinforce the quality of pseudo-labels and then improve the performance of the relation classification module.
Specifically, in a batch-mode training process, we use the up-to-date learned soft matcher to assign all unmatched sentences with weighted labels, which further serve as an auxiliary learning objective for the relation classifier module.

Extensive experiments on two public datasets, TACRED~\cite{zhang2017position} and Semeval~\cite{hendrickx2009semeval}, demonstrate that the proposed \NERO consistently outperforms baseline methods  using the same resources by a large margin.
To investigate the label efficiency of \NERO, 
we further conduct a user study to compare models trained with rules and labeled sentences, respectively, both created under the same time constraint.
Results show that to achieve similar performance, \NERO requires about ten times fewer human efforts in annotation time than traditional methods of gathering labels.

The \textbf{contributions} of our work are summarized as follows:
(1) Our proposed method, \NERO, is among the first methods that can learn to generalize labeling rules for training an effective relation extraction model, without relying on additional sentence-level labeled data;
(2) We propose a learnable rule matcher that can semantically ground rules for sentences with similar meaning that are not able to be hard-matched. This method also provides interpretable predictions and supports few-shot learning ability on unseen relations;
(3) We conduct extensive experiments to demonstrate the effectiveness of our framework in a low-resource setting and show careful investigations on the efficiency of using human efforts.


\section{Problem Formulation}
\label{sec:background}
We first introduce basic concepts and their notations and then present the problem definition as well as the scope of the work.

\smallskip
\noindent
\textbf{Relation Extraction.}
Given a pair of entity strings ($e_\text{subj}$, $e_\text{obj}$), identified as the \texttt{subject} and \texttt{object} entities respectively, the task of relation extraction (RE) is to predict (classify) the relation $r \in \mathcal{R} \cup \{\textsc{None}\}$ between them, where $\mathcal{R}$ is a pre-defined set of relations of interest. 
Specifically, here we focus on \textit{sentence-level} RE~\cite{hendrickx2009semeval,zhang2017position,lin2019dualre}, which aims to predict the relation of entity mention pairs in a sentence $s$---\ie, identifying the label $r \in \mathcal{R}$ for an \textit{instance} in the form of $(e_\text{subj}, e_\text{obj}; s)$\footnote{We use ``instance" and "sentence" exchangeably in the rest of the paper.} and yielding the prediction $(e_\text{subj}, r, e_\text{obj}; s)$. 


\smallskip
\noindent
\textbf{Labeling Rules.}
As an alternative to instance-level human-annotated data (\eg, instance-label pairs), labeling rules formalize human's domain knowledge in a structured way, and can be either directly used for making predictions over test instances~\cite{agichtein2000snowball}, or applied to generate labeled instances for model learning~\cite{ratner2016data}. Formally, an inductive labeling rule consists of a \textit{rule body} and a \textit{rule head}. 
In the context of relation extraction, we consider the cases where the rule body is a textual pattern $p = [\textsc{subj-type}; c; \textsc{obj-type}$], in which $c$ denotes a word sequence between the two entities in a sentence (\ie, context), and \textsc{subj-type} and \textsc{obj-type} specify the entity types of the subject and object entities required by the rule body.
An instance $(e_\text{subj}, e_\text{obj}; s)$ is \textit{hard-matched} by a rule $p$ if and only if $p$'s context $c$ exactly match the context between $e_\text{subj}$ and $e_\text{obj}$ in $s$  
and the entity types of $e_\text{subj}$ and $e_\text{obj}$ match $p$'s associated entity types.
For example, sentence $s_1$ in Fig. \ref{fig:rule_eg} is hard-matched by rule $p_3$.

Such labeling rules can either be hand-written by domain experts, or automatically generated from text corpora with a knowledge base that shares the relations of interests. 
In this work, we adopt a hybrid approach (as detailed in Sec.~\ref{sec::rule_labeling}): 
1) we first extract surface patterns (\eg frequent patterns, relational phrases, \etc) as candidate rule bodies from raw corpora (after running named entity recognition tool) using pattern mining tools, and 2) ask human annotators to assign relation labels (\ie, rule heads) to the rule bodies.
Then, we apply these human-annotated rules over instances from a raw corpus to generate labeled instances (as training data). 
Compared to directly annotating instances, our study will show that annotating rules is much more label-efficient, as one rule may match many instances in the corpus with reliable confidence, generating multiple labeled instances.
Moreover, annotating rules may require similar (or even less) time than annotating instances since they are shorter in length, as shown in our later  analysis.
\begin{thm:eg}[Labeling Rule Generation]\label{ex:problem}
Given instances such as \{(Bill Gates, Microsoft; ``Bill Gates founded Microsoft.''), (Bill Gates, Steve Jobs; ``Guests include Bill Gates and Steve Jobs.'')\}, the candidate patterns include \{``\textsc{subj-person} founded  \textsc{obj-person}'', ``\textsc{subj-person}, \textsc{obj-person}''\}
, and the rule set after human annotation includes \{``\textsc{subj-person} founded  \textsc{obj-person}'' $\rightarrow$ \textsc{founded\_by}\}. The second pattern is filtered out by annotators due to its low quality.
\end{thm:eg}

\noindent 
\textbf{Problem Definition.} 
In this paper, we focus on sentence-level relation extraction using labeling rules created by a semi-automatic method. Specifically, given a raw corpus consisting of a set of instances \begin{small}$\mathcal{S} = \{(e_\text{subj}^i, e_\text{obj}^i; s^i)\}_{i=1}^N$\end{small}, 
our approach first extracts surface patterns \begin{small}$\{[\textsc{subj-type}; c; \textsc{obj-type}]\}$\end{small} as candidate rule bodies, and then annotates them to generate labeling rules
\begin{small}$\mathcal{P}=\{p^k\}_{k=1}^{|\mathcal{P}|}$\end{small} with $p^k$ denoting $[\textsc{subj-type}^k; c^{k}; \textsc{obj-type}^{k}] \rightarrow r_k$ and $r_k \in \mathcal{R} \cup \{\textsc{None}\}$.
Next, our framework aims to leverage both the rules $\mathcal{P}$ and the corpus $\mathcal{S}$ to learn an effective relation classifier \begin{small}$f: \mathcal{S} \rightarrow \mathcal{R} \cup \{\textsc{None}\}$\end{small} which can predict relation for new instances in unseen sentences.

In addition to predicting relations already observed in the training phase, we aim to  \textbf{predict unseen relations} solely based on new rules specified for the unseen relations at the test time---\ie, a few-shot learning scenario. Such a generalization ability is desirable for domain adaptation. In this setting, the model is trained on a corpus $\mathcal{S}$ using labeling rules defined on relation set $\mathcal{R}_c$. 
In the testing phase, the model is expected to make predictions on unseen relation types $\mathcal{R}_u$ given some corresponding labeling rules.

\smallskip
\noindent
\textbf{Our Focus.}
Instead of relying on pre-existing knowledge bases to conduct distant supervision, we focus on the scenarios where labeling rules are relatively cheaper and faster to obtain by automatically inducing from large text corpora (as compared to manually curating a KB from scratch).
We study neural rule grounding to integrate information from rules and corpora into learning an effective relation extraction model that can generalize better than solely using rules in a hard-matching manner.
We mainly use surface pattern rules (\eg, similar to the ones used in \cite{roth2014effective}) and leave the study of more complex rules (\eg, regular expression) as future work. 
\begin{figure*}[!t]
    \vspace{-0.1cm}
    \centering
    \includegraphics[width=0.97\linewidth]{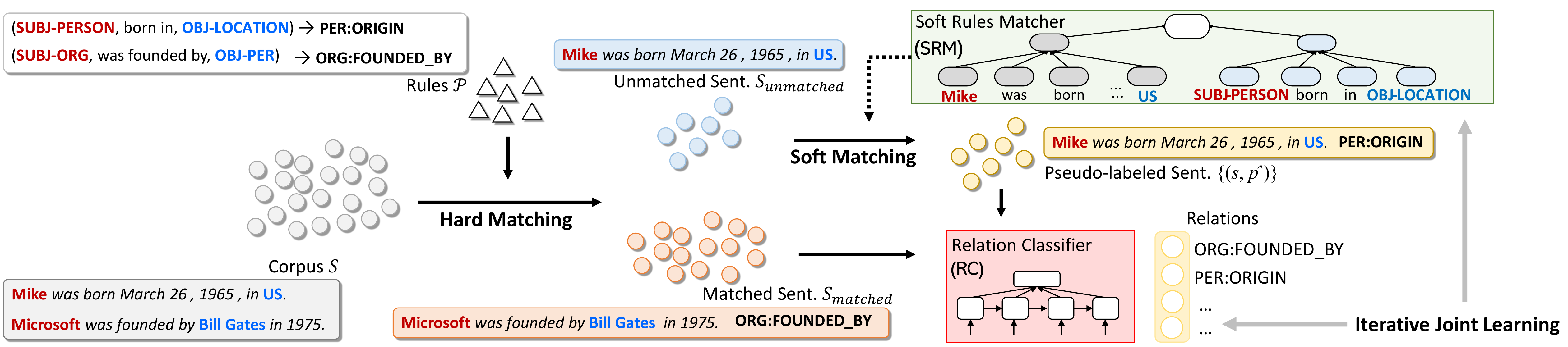} 
     \vspace{-0.3cm}
    \caption{\textbf{Overview of the \NERO framework.} Each unmatched sentence is first annotated by the soft rule matcher (\texttt{SRM}) to generate pseudo labels, and then fed into the relation classifier (\texttt{RC}) to update the model parameters. The whole framework is trained iteratively and jointly, based on multiple loss functions as introduced in Sec.~\ref{sec::joint_learning}. 
    }
    \label{fig::frameoverview}
     \vspace{-0.3cm}
\end{figure*}

\section{Neural Rule Grounding (\NERO)}
\label{sec::framework}
\wx{This section introduces the important concepts in \NERO, and provides details of the framework design and learning objectives.
We first present our key motivation, then give an overview of \NERO framework (Sec.~\ref{ssec:overview}) and introduce the details of each component (Secs.~\ref{sec::rule_labeling}-\ref{sec::rule_matcher_module}).
Lastly, we show how we jointly learn model parameters in \NERO for sequence encoding and rule matching (Sec.~\ref{sec::joint_learning}).

\smallskip
\noindent
\textbf{Motivation of Soft Rule Matching.}
Traditional methods of using labeling rules for relation extraction are mainly based on exact string/pattern matching (\ie, \textit{hard-matching}), where a sentence can be annotated by a rule if and only if it has exactly the same surface form as the rule body.
While being easy to implement and yielding relatively high precision for the prediction, these hard-matching methods only achieve sub-optimal performance because of the severe low recall issue (see Fig. \ref{fig:rule_eg} for an example).
Such a low coverage of using rules can only produce a minimal number of ``labeled" sentences for learning a RE model, which may be far from enough for data-hungry neural models.
To address this problem, we would like to study ``soft-matching'' for exploiting the rules for relation extraction.
A soft-rule matcher is proposed to semantically match a rule even when the surfaces are different (e.g. ``founded'' v.s. ``created'').
Towards improving relation extraction with soft-matching, we would like to learn a \texttt{SRM} such that we can 
assign \textit{pseudo-label}s to sentences.
Further, we can use these pseudo-labeled sentences to learn a sentence-level relation classifier.

}

\vspace{-0.1cm}
\subsection{Framework Overview}
\label{ssec:overview}
The \NERO framework consists of a relation classifier (\texttt{RC}) and a soft rule matcher (\texttt{SRM}). 
The \texttt{RC} is for learning neural representations for sequences of words, from either an instance or a rule body (Sec.~\ref{sec::sequence_encoder_module}).
The \texttt{SRM} gives a matching score to indicate the similarity between a rule body and an instance (Sec.~\ref{sec::rule_matcher_module}). 
Previous works \cite{batista2015semi} mostly adopt a \textit{fixed} metric function (\eg, cosine similarity measure between the sequence embeddings).
In contrast, our \texttt{SRM} is a \textit{learnable} neural network and thus can be more effective in modeling the semantic similarity between sentences and rule bodies.
As shown in Fig.~\ref{fig::frameoverview},
we first apply our collected rules to a raw corpus $\mathcal{S}$ to perform \textit{hard matching}.
$\mathcal{S}$ will then be divided into two subsets: ``\textit{hard-matched sentences}'' $\mathcal{S}_\text{matched}$ and ``\textit{unmatched sentences}'' $\mathcal{S}_\text{unmatched}$.
Lastly, we use \texttt{SRM} and rules to iteratively generate pseudo labels over $\mathcal{S}_\text{unmatched}$, while jointly learning \texttt{RC} and \texttt{SRM} with the pseudo labels to promote a common representation space (Sec.~\ref{sec::joint_learning}).


\subsection{Labeling Rule Generation}
\label{sec::rule_labeling}
As introduced in Sec.~\ref{sec:background}, our candidate rule bodies (\ie, surface patterns) are automatically extracted from the raw corpus.
In this work, we adopt a simple yet effective pattern mining method. Given a raw corpus, we first replace entities with entity type masks \textsc{SUBJ/OBJ-NER}, where ``NER" denotes entity type (same as the procedure in \cite{zhang2017position}).
Then we pick the word sequences between and including the two entities as candidate rules. So each candidate rule is just a short phrase / sentence containing two masked entities. To reduce the annotation efforts and ensure the popularity of rules, we convert all words to their root form using Porter stemming algorithm \cite{willett2006porter} and only keep the rules whose stemmed form appears at least $N$ times in the whole corpus. Finally, we ask human annotators to select rules that indicate a relation and assign labels to them. This process is similar to label a sentence with a pre-defined relation or \textsc{None}. In spite of the existence of more complex rule extraction methods such as shortest dependency path \cite{nakashole2012patty, qu2018weakly} and meta patterns \cite{jiang2017metapad}, we adopt frequent word sequences in this work because of their simplicity and high readability. Our framework can also be adapted to other rule generation methods as long as they can generate human-readable candidate rules.
Regardless of their high generalization ability, labeling rules may bias the label distribution, \eg \textsc{per:title} counts for $19\%$ in the official training data while only counts for $5.5\%$ in the matched sentences. This domain gap poses another challenge for learning with rules.

\subsection{Relation Classifier (\texttt{RC})}
\label{sec::sequence_encoder_module}
The relation classifier (named as \texttt{RC}) is the end product of our \NERO framework, which can
represent a relation instance $(e_1, e_2, s)$
into vector embeddings.
This module is general and can use various designs (\eg, attention CNN \cite{wang-etal-2016-relation}, attention RNN \cite{zhou-etal-2016-attention, zhang2017position}, and GCN \cite{zhang2018graph, guo2019attention}).
In this work, we stick with the LSTM + ATT model \cite{zhou-etal-2016-attention}.
Given a sequence of $n$ words, we first look up their word embeddings as $\{\mathbf{x_t}\}_{t=1}^n$. 
Then we use a bi-directional LSTM network \cite{hochreiter1997long} to obtain the contextualized embeddings of each word $\{\mathbf{h_t}\}_{t=1}^n$ where $\mathbf{h_t}\in\mathbb{R}^{d_h}$.
Finally, we employ an attention layer~\cite{bahdanau2014neural} to get a sentence representation $\bc$:
\begin{align}
    \label{eq::attention-1}
       \{\mathbf{h_t}\}_{t=1}^n &= \textrm{BiLSTM}\left(\{\mathbf{x_t}\}_{t=1}^n\right)\\
    \alpha_t &= \frac{\textrm{exp} (\mathbf{v}^{\text{T}}~\textrm{tanh}(\mathbf{A} \mathbf{h_t}))}{\sum_{t'=1}^n \textrm{exp} (\mathbf{v}^{\text{T}}~\textrm{tanh}(\mathbf{A} \mathbf{h_{t'}}))} \\
    \label{eq::attention-3}
    \bc &= \sum_{t=1}^n \alpha_t \bh_t
\end{align}
where $\mathbf{A} \,{\in}\, \mathbb{R}^{d_a \times d_h}$, $\bv \,{\in}\ \mathbb{R}^{d_a}$ are learnable model parameters for the attention mechanism.
We can get the relation type probability distribution by feeding $\mathbf{W_\text{rc}}\bc$ into a SoftMax layer:
$$\texttt{RC}(s, e_{\text{subj}},e_{\text{obj}}) = \textrm{SoftMax}(\mathbf{W_{\textrm{rc}}}\mathbf{c}),$$
where $\mathbf{W_\text{rc}}\in\mathbb{R}^{d_h \times |\mathcal{R}|}$ is a matrix to learn.
In other words, we are using the output vector of the $RC$ for modeling the conditional distribution of relation  $\prob_{\theta_{RC}}(r=i|s)=\texttt{RC}(s,e_{\text{subj}},e_{\text{obj}})[i]$, meaning that $i-th$ element of the \texttt{RC} output vector as the probability of the $i$-th relation for the input sentence $s$.
The $\theta_\text{rc}$ denotes the set of model parameters of \texttt{RC} that we need to learn, including $\mathbf{A}, \mathbf{v}, \mathbf{W_\text{rc}}$, and the weights of the BiLSTM.




\subsection{Soft Rule Matcher (\texttt{SRM})}
\label{sec::rule_matcher_module}
\begin{figure}[!tb]
\vspace{-0.4cm}
    \centering
    \includegraphics[width=0.95\linewidth]{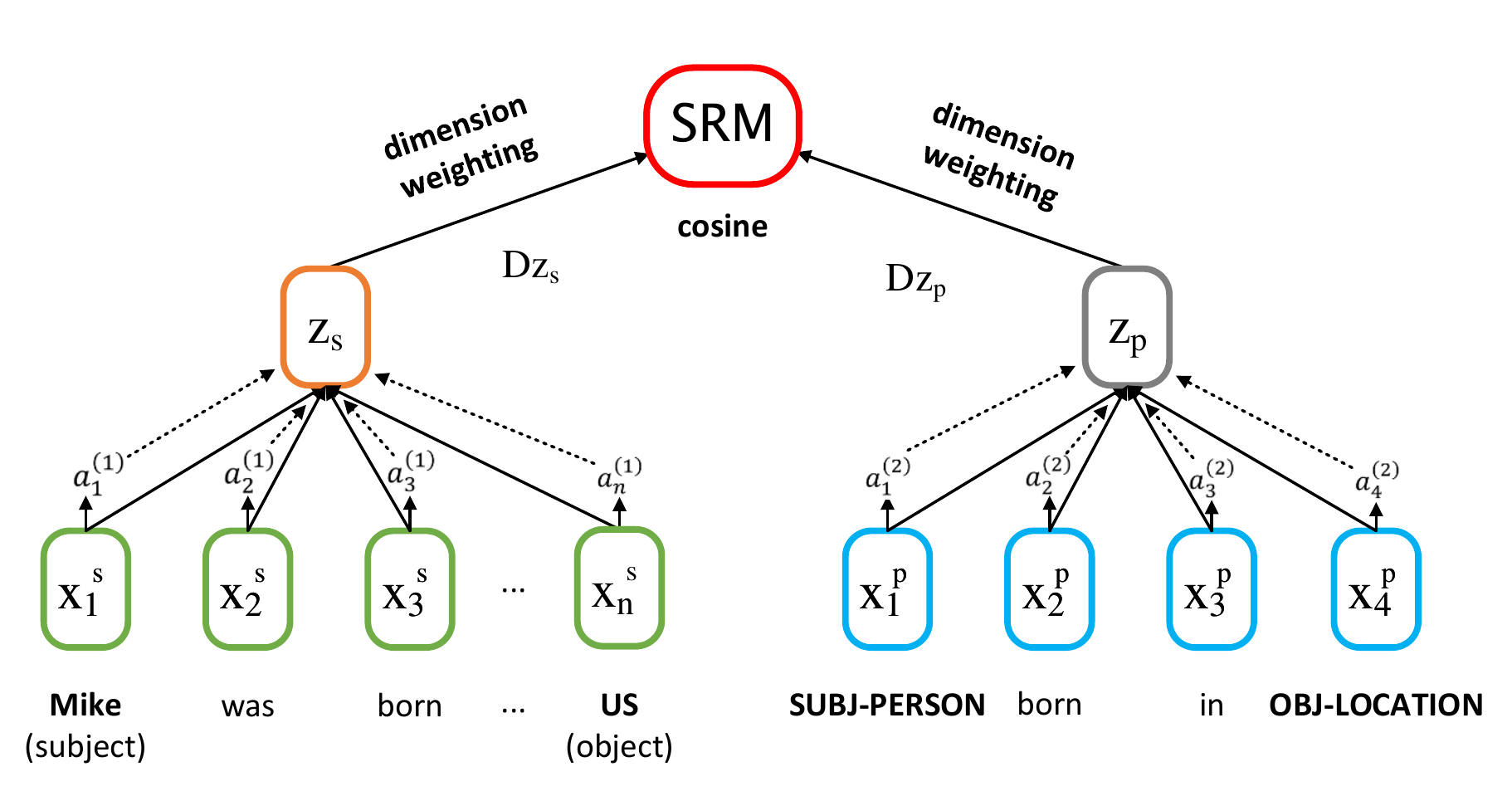}
     \vspace{-0.4cm}
    \caption{\textbf{Detailed architecture of the soft rule matcher (\texttt{SRM}).} The cosine similarity between two embeddings indicates the degree of matching between rules and sentences.}
    \label{fig::softmatch}
    \vspace{-0.2cm}
\end{figure}

The core component of the \NERO is the soft rule matcher (named to be \texttt{SRM}), which is defined as a neural function for modeling the matching score between a sentence $s$ and a rule pattern $p$. 
We formulate the Soft Rule Matcher as a function $\texttt{SRM}:(s, p) \rightarrow [-1, 1]$. 
As shown in Fig. \ref{fig::softmatch}, we first map the sentence $s$ and rule $p$ into the same embedding space by the \texttt{RC}, and then apply a distance metric on their embeddings to get the matching score ranging from $-1$ to $1$.
Note that a matching score of 1 indicates the sentence can be hard-matched by the rule.
Because the rules are typically short phrases, we use word-level attention -- a weighted continuous bag-of-words model to calculate matching scores.
We explored more designs in later experiments and this part can be further studied in future work.
We represent a sentence and a rule pattern as the sequence of the word embeddings, $\{\mathbf{x^s_t}\}_{t=1}^n$ and  $\{\mathbf{x^p_t}\}_{t=1}^m$, respectively.
By applying the attention on word embeddings instead of contextualized embedding generated by LSTM, 
we aim to reduce over-fitting given the scarcity of rules per relation.
Note that we are using a different set of parameters for this word embedding-level attention ($\mathbf{B}$ and $\mathbf{v}$).
Specifically, we have 
\begin{align}
    \mathbf{z_s} &= \sum_{t=1}^n \frac{\textrm{exp} (\mathbf{u}^{\text{T}}~\textrm{tanh}(\mathbf{B} \mathbf{x^s_t}))}{\sum_{t'=1}^n \textrm{exp} (\mathbf{u}^{\text{T}}~\textrm{tanh}(\mathbf{B} \mathbf{x^s_{t'}}))} \mathbf{x^s_t}, \\
    \mathbf{z_p} &= \sum_{t=1}^m \frac{\textrm{exp} (\mathbf{u}^{\text{T}}~\textrm{tanh}(\mathbf{B} \mathbf{x^p_t}))}{\sum_{t'=1}^m \textrm{exp} (\mathbf{u}^{\text{T}}~\textrm{tanh}(\mathbf{B} \mathbf{x^p_{t'}}))} \mathbf{x^p_t}, \\
    \texttt{SRM}(s, p) &= \frac{ (\mathbf{D}\mathbf{z_s})^T (\mathbf{D} \mathbf{z_p})}{\|\mathbf{D} \mathbf{z_s}  \| \| \mathbf{D}\mathbf{z_p}  \|},
\end{align}
where $\mathbf{z_s}$, $\mathbf{z_p}$ are the representations for the sentence $s$ and the rule $p$ respectively, and $\mathbf{D}$ is a trainable diagonal matrix which denotes the importance of each dimension.
Similar idea has been shown effective in other tasks for capturing semantically close phrases, and it generalizes well when only limited training data is available~\cite{yu2015learning}.
In practice, a sentence may be very long and contain much irrelevant information. 
So for the sentence, we only keep the words between (and including) the subject and object entities, which is a common practice in other rule-based RE systems \cite{batista2015semi, qu2018weakly}.


\subsection{Joint Module Learning}
\label{sec::joint_learning}


We now formulate the overall learning process of \NERO in a low-resource learning scenario, where we only have collected rules but no sentence-level labels at all.
Given a raw corpus $\mathcal{S}$ (i.e. a large set of sentences without human-annotated labels) and a set of inductive labeling rules $\mathcal{P}$, 
we first exactly match every rule on $\mathcal{S}$ and get a set of \textbf{hard-matched sentences} named $\mathcal{S}_\text{matched}$. 
These hard-matched labels are pretty accurate\footnote{We conduct an experimental analysis on the quality of these matches in Sec.~\ref{sec:exp}.}, which can serve the purpose as training examples. 
The other \textbf{unmatched sentences}, denoted as $\mathcal{S}_\text{unmatched}$ are also informative for our model, while they are ignored by most prior works. 
Although not able to be exactly matched by any rules, 
the unmatched sentences can semantically align the meaning of some rules and thus help improve the RE performance.
By applying the \texttt{SRM} upon them to create pseudo-labels, we can further utilize the hidden useful information as the supervision for learning the final RE model. 
Apart from that, the \texttt{SRM} is also optimized for improving the quality of matching in the meantime. 
Finally, supervision signals from $\mathcal{S}_\text{matched}$, $\mathcal{S}_\text{unmatched}$, and $\mathcal{P}$ together train the RE model in a joint learning schema.

\medskip
\noindent \textbf{Learning with Hard-Matched Sentences ($\mathcal{S}_\text{matched}$)}. 
As the primary goal of the learning, we want to minimize the error of the \texttt{RC} in classifying relations in the sentences in $\mathcal{S}_\text{matched}$ respect to their matched relation labels.
Given a hard-matched sentence $s\in \mathcal{S}_\text{matched}$ and its associated relation $r_s \in \mathcal{R}$, 
we aim to minimize the cross-entropy loss ${L}_{\text{matched}}$ as follows.
\begin{equation}
\label{eq::annotated_loss}
    {L}_{\text{matched}} (\theta_{RC}) = \mathbb{E}_{s \sim \mathcal{S}_\text{matched}} \left[-\log{\prob_{\theta_{RC}}(r = r_s | s)} \right].
\end{equation}

\medskip
\noindent 
\textbf{Learning with Rules ($\mathcal{P}$)}. 
To incorporate our collected rules for exploiting the unmatched sentences $\mathcal{S}_\texttt{unmatched}$ as well, 
we propose two additional auxiliary tasks for imposing rules in the \texttt{RC} and \texttt{SRM}.
First,
we can treat the rule body $p\in\mathcal{P}$ as a ``sentence'' and its rule head $r_p$ as its associated label,
thus forming a labeled instance $(p, r_p)$.
With such rule-reformatted instances as training data, we aim to minimize the error for \texttt{RC} in classifying them as follows.
This objective helps \texttt{RC} to explicitly reinforce the memory of \texttt{RC} about the collected rules.
\begin{equation}
\label{eq::rule_loss}
    {L}_\text{rules} (\theta_{RC})= \mathbb{E}_{p \sim \mathcal{P}} \left[-\log{\prob_{\theta_{RC}}(r=r_p | p)}\right]
\end{equation}

More importantly, we present how we utilize the rules for learning the soft rule matcher (\texttt{SRM}) such that we can exploit the unmatched sentences as distant supervision.
Our key motivation here is that a good matcher should be able to clustering rules with the same type, and thus we expect the \texttt{SRM} to increase the distances between rules with different types and reduce the variance of the distances between rules  with the same types.
Simply put, given two rules, their similarity score should be high if they belong to the same relation type and low otherwise. 

We use the contrastive loss \cite{Neculoiu2016LearningTS} for this objective. 
Given a rule $p\in \mathcal{P}$ and its relation $r_p\in \mathcal{R}$, we divide all the other rules with the same relation type as $\mathcal{P}_+(p)=\{p'| r_{p'} = r_p \}$, and the ones with different types as $\mathcal{P}_-(p)=\{p'| r_{p'} \neq r_p \}$. 
The contrastive loss aims to pull the rule $p$ closer to its most dissimilar rule in $\mathcal{P}_+(p)$ and in the meantime push it away from its most similar rule in $\mathcal{P}_-(p)$. Mathematically, the loss is defined as follows:
\begin{equation}
    {L}_\text{clus} = \mathbb{E}_{p \sim \mathcal{P}} \left[ \max_{p_i\in \mathcal{P}_+(p)}{\text{dist}_+(p, p_i)} - \min_{p_j \in \mathcal{P}_-(p)}{\text{dist}_-(p, p_j)} \right],
\end{equation}
where the measurement of the most dissimilar same-type rule and most similar different-type rule can be measured with the distances defined as below:
\begin{align*}
    \text{dist}_+(p, p_i) &= \max \big(\tau - \texttt{SRM}(p, p_i), ~0 \big)^2, \\
    \text{dist}_-(p, p_j) &= 1 - \max \big( \texttt{SRM}(p, p_j), ~0 \big)^2.
\end{align*}
$\tau$ is a hyper-parameter for avoiding collapse of the rules' representations. 
Minimizing ${L}_\text{clus}$  pushes the matching scores of rules that are of the same relation type up to $\tau$ and pull the matching score down to $0$ otherwise.
Without labels about the alignment between rules $\mathcal{P}$ and unmatched sentences $\mathcal{S}_\text{unmatched}$, this objective can be seen as a secondary supervision for \texttt{SRM} to train the parameters $\theta_{\texttt{SRM}}$ using $\mathcal{P}$ itself only by this contrastive loss.
\vspace{2mm}

\begin{figure}[!t]
\vspace{-0.2cm}
    \centering
    \includegraphics[width=1.01\linewidth]{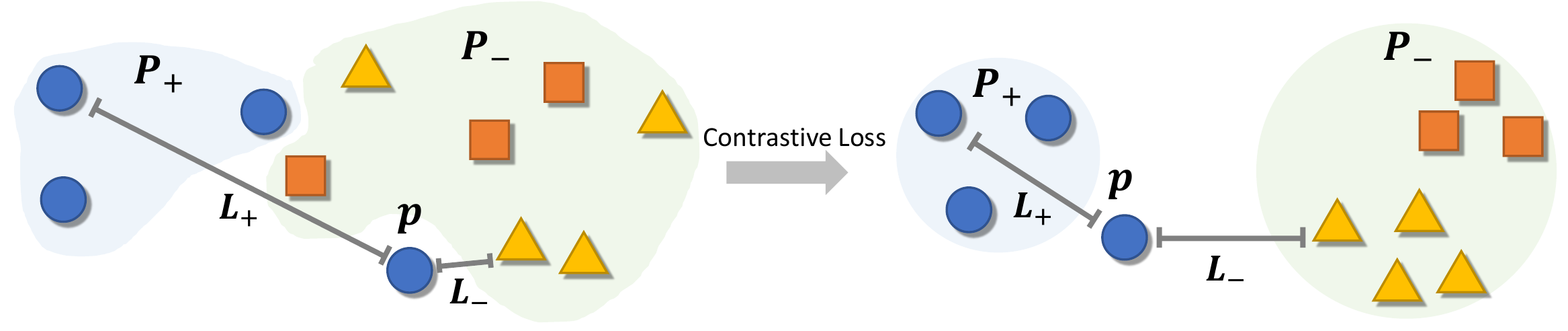}
    \vspace{-0.6cm}
    \caption{\textbf{The contrastive loss in learning the \texttt{SRM}}, which increases the matching score of rules with the same relation type and decrease the matching score otherwise. 
    }
    \label{fig::rule_cluster}
     \vspace{-0.2cm}
\end{figure}

\medskip
\noindent 
\textbf{Learning with Unmatched Sentences ($\mathcal{S}_\text{unmatched}$)}. We use the \texttt{SRM} to label unmatched sentences $\mathcal{S}_\text{unmatched}$ (see Fig. \ref{fig::frameoverview}). 
As the labels generated by the \texttt{SRM} may be noisy, 
we propose to use pseudo-labeling \cite{lee2013pseudo} with instance weighting \cite{jiang2007instance} to alleviate the noise.
Each unmatched sentence is weighed by the matching score from the \texttt{SRM}. 
Specifically, for each unmatched sentence $s \in \mathcal{S}_\text{unmatched}$, we first apply the \texttt{SRM} to compute its matching scores with each rule $p\in\mathcal{P}$ and then assign the pseudo-label to the sentence $s$ as the relation of the highest-score rule $r_{\hat{p}}$, where $$\hat{p}={\arg\max_{p\in\mathcal{P}}\texttt{SRM}(s, p)}.$$ The corresponding weight for a particular pseudo-labeled instance, $(s_i, \hat{p}_i)$, is the \texttt{SoftMax} over all the sentences (in a mini-batch when training; see Sec.~\ref{sec::learning} for more details):
\begin{equation}
w_{s} = \frac{\textrm{exp}(\sigma \texttt{SRM}(s, \hat{p}_i))}{\sum_{s'\in \mathcal{S}_\text{unmatched}} \textrm{exp}(\sigma \texttt{SRM}(s', \hat{p}_j))},
\end{equation}
where $\sigma$ represents the temperature of the SoftMax function. The loss function for each batch of unmatched sentences would be:
\begin{equation}
    {L}_{\text{unmatched}} (\theta_{RC}) = \mathbb{E}_{s \sim \mathcal{S}_\text{unmatched}} \left[-w_s\log{\prob_{\theta_{RC}}(r = r_{\hat{p}} | s)} \right].
\end{equation}

\medskip
\noindent 
\textbf{Joint Optimization Objective}.
The whole framework is jointly trained under the overall loss function:
\begin{equation*}
\label{eq.l}
    \begin{aligned}
    {L} (\theta_{\text{RC}}, \theta_{\text{SRM}}) &= {L}_\text{matched}(\theta_{\text{RC}}) ~+ ~\\ 
     &\alpha\cdot {L}_\text{rules}(\theta_{\text{RC}})  +  \beta\cdot  {L}_\text{clus}(\theta_{\text{SRM}}) + \gamma\cdot {L}_\text{unmatched}(\theta_{\text{RC}}).
\end{aligned}
\end{equation*}
To sum up, the proposed \NERO framework has two trainable components: a relation classifier (\texttt{RC}) and a soft rule matcher (\texttt{SRM}) with their specific parameters ($\theta_{\text{RC}}$ and $\theta_{\text{SRM}}$).
Our primary task is to minimize the error of the \texttt{RC} on the hard-matched sentences (${L}_\text{matched}$).
To exploit the unmatched sentences with collected rules, we first let \texttt{RC} to explicitly learn the rules with the objective of rule classification (${L}_\text{rules}$) and also learn the \texttt{SRM} with the help of contrastive loss (${L}_\text{clus}$) for clustering rules.
We jointly learn the two modules by connecting them through pseudo-labeling on the unmatched sentences with the \texttt{SRM} and then expect the \texttt{RC} has better performance on such pseudo-labeled unmatched sentences as well (${L}_\text{unmatched}$).

\begin{algorithm}[!t]
    \KwInput{A raw corpus $\mathcal{S}$, pre-defined relations $\mathcal{R}\cup \{\textsc{None}\}$.}
    \KwOutput{A relation classifier $f: \mathcal{S} \rightarrow \mathcal{R}\cup \{\textsc{None}\}$.}
    Extract candidate rules from $\mathcal{S}$ with pattern mining tools. \\
    Ask human annotators to select and label the candidate rules to get $\mathcal{P}$. \\
    Partition $\mathcal{S}$ into $\mathcal{S}_\text{matched}$ and $\mathcal{S}_\text{unmatched}$ by hard-matching with $\mathcal{P}$.
    \caption{Optimization of \NERO model}
    \While{$L$ in \eq \ref{eq.l} not converge}{
        Sample batch $\mathcal{B}_m = \{(s_i, r_i)\}_{i=1}^n$ from $\mathcal{S}_\text{matched}$. \\
        Update $L_\text{matched}$ by \eq \ref{eq::annotated_loss}. \\
        Sample batch $\mathcal{B}_u=\{s_j\}_{j=1}^m$ from $\mathcal{S}_\text{unmatched}$. \\
        \ForEach{$s \in \mathcal{B}_u$}{
            Find highest-scored rule $\hat{p}$ and pseudo label $r_{\hat{p}}$ by \texttt{SRM}. \\
        } 
        Update $L_\text{unmatched}$ by \eq \ref{eq::batch_unmatched}.\\
        Update $L_\text{rules}$ by \eq \ref{eq::rule_loss}. \\
        \ForEach {$p\in \mathcal{P}$}{
            Calculate $\texttt{SRM}\,(p, p')$ for each $p'\in \mathcal{P}-\{p\}$.\\
            Update $L_\text{clus}$. \\
        }
        $L = L_\text{matched} + \alpha \cdot L_\text{rules} + \beta \cdot L_\text{clus} + \gamma \cdot L_\text{unmatched} $.\;
        Update model parameters \emph{w.r.t.} $L$.
    }
\label{algo::NERO}
\end{algorithm}

\section{Model Learning and Inference}
\label{sec:modellabel}
In this section, we introduce the specific training details of our proposed framework, and how the model conduct inference.
\subsection{Parameter Learning of \NERO}
\label{sec::learning}
\NERO starts with a raw corpus $\mathcal{S}$ and pre-defined relations $\mathcal{R} \cup \{\textsc{None}\}$.
It first extracts candidate rules from $\mathcal{S}$ with pattern mining tools, then asks human annotators to select and label the rules to get the rule set $\mathcal{P}$.
Before training, \NERO does a quick pass of $\mathcal{S}$ by hard-matching to split it into hard-matched sentences $\mathcal{S}_\text{matched}$ and unmatched sentences $\mathcal{S}_\text{unmatched}$. Both of them are utilized to train the relation classifier \texttt{RC} in a joint-training manner. Algorithm \ref{algo::NERO} summarizes the overall training procedure.

The joint training objective is efficiently optimized by batch-mode training. 
Specifically, in each training step we randomly sample a batch of hard-matched sentences $\mathcal{B}_{\text{m}}$ from $\mathcal{S}_{\text{matched}}$, and then calculate the average cross-entropy loss in the batch by ${L}_\text{matched}$.
Similarly, we sample $\mathcal{B}_{\text{u}}$ from $\mathcal{S}_{\text{unmatched}}$, and generate the pseudo-labels with the entire rule set $\mathcal{P}$, and calculate normalized weights for ${L}_\text{unmatched}$ by:
\begin{equation}
\label{eq::batch_unmatched}
    \begin{gathered}
    w_{s} = \frac{\textrm{exp}(\sigma \texttt{SRM}(s, \hat{p}_i))}{\sum_{s'\in \mathcal{B}_{\text{u}}} \textrm{exp}(\sigma \texttt{SRM}(s', \hat{p}_j))}, \\
    {L}_{\text{unmatched}} (\theta_{RC}) = \frac{1}{\left|\mathcal{B}_\text{u}\right|}\sum_{s \in \mathcal{B}_\text{u}} \left[-w_s\log{\prob_{\theta_{RC}}(r = r_{\hat{p}} | s)} \right].
\end{gathered}
\end{equation}
The normalized instance weights ensure the scale of ${L}_{\text{unmatched}}$ to be stable across different steps. Finally, we calculate $L_\text{rules}$ and $L_\text{clus}$ with the entire $\mathcal{P}$.
For $L_\text{rules}$, the loss is averaged for all $p\in \mathcal{P}$.
For $L_\text{clus}$, we take all $p\in \mathcal{P}$ to calculate the contrastive loss w.r.t. $\mathcal{P} - \{p\}$, and average the losses.
We do not sample $\mathcal{P}$ due to the relatively small number of rules and simple structure of the soft rule matcher (\texttt{SRM}).
We stop training when the joint training loss ${L}$ converges and take \texttt{RC} as the output.
Note that word vectors are also trainable parameters that are initialized by pre-trained embeddings.

\subsection{Model Inference}
\label{sec::infer}

Model inference in \NERO aims at predicting the relation of a new sentence. Intuitively, we can give out a prediction by passing the sentence through the \texttt{RC} and return the relation with the highest probability as the prediction. As an alternative, we can also predict the relation using the \texttt{SRM}. Given a sentence, we can first find the most similar rule and return its rule head as the prediction. This alternative way of inference can be applied for predicting \textit{unseen relations} given new rules in the testing time. In experiments, the first method shows much better performance, since \texttt{RC} can capture rich contextual information while the \texttt{SRM} cannot.

For predicting ``\textsc{None}'' examples,
\textit{\textbf{}}our model filters out the predictions that our model is least certain\footnote{A threshold $\delta$ tuned on dev set is used.} about. Specifically, we measure the uncertainty using the entropy of softmax distribution or the similarity score produced by \texttt{SRM}.

\section{Experiments}
\label{sec:exp}
In this section,
we first introduce the datasets and compared baseline methods.
Then, we illustrate the detailed setup and present extensive experiments with discussion and analysis.
Finally, we conduct user studies to investigate the efficiency of our proposed approach.

\begin{table}[t]
\vspace{-0.2cm}
    \centering
    \scalebox{0.75}{
    \begin{tabular}{ccccccc}
    \toprule
         Dataset& \# Train / Dev / Test& \# Relations&\# Rules& \# matched Sent.  \\
    \midrule
        TACRED \cite{zhang2017position}& 75,049 /25,763 / 18,659& 42& 270& 1,630 \\
        SemEval \cite{hendrickx2009semeval}& 7,199 /800 / 1,864& 19& 164& 1,454 \\
        \bottomrule \\
    \end{tabular}}
    \vspace{-0.1cm}
    \caption{Statistics for TACRED and SemEval datasets.}
    \label{tab:data_statistics}
    \vspace{-0.6cm}
\end{table}

\subsection{Data Preparation}
We choose two public and widely-used sentence-level relation extraction datasets in our experiments (see Tab.~\ref{tab:data_statistics}) as follows. For both datasets, we construct the rule set as illustrated in Section \ref{sec::rule_labeling}\footnote{We filter candidate patterns by requiring frequency $\ge 3$.}.

\begin{itemize}
    \item \textbf{TACRED} \cite{zhang2017position} (TAC relation extraction dataset) contains more than 100,000 sentences categorized into 42 relation types. 
Among the sentences, 79.5\% of the examples are labeled as \textsc{None}. 
We construct 270 rules which 1,630 hard-matched sentences in the official training data; 
    \item \textbf{SemEval} 2010 Task 8 \cite{hendrickx2009semeval} contains about 10,000 sentences with 19 relation types, where 17.4\% of the sentences are \textsc{None}. 
    We construct 164 rules which hard-match 1,454 instances in the official training data.
\end{itemize}

\subsection{Compared Methods}
Recall Sec.~\ref{sec::joint_learning}, we first apply hard-matching on the unlabeled sentences and partition them into hard-matched sentences ($\mathcal{S}_\text{matched}$) and unmatched sentences ($\mathcal{S}_\text{unmatched}$). 
Our framework is applicable in both training with $\mathcal{S}_\text{matched}$ and training with $\mathcal{S}_\text{matched} \cup \mathcal{S}_\text{unmatched}$.
Thus,
we compare our models with \textit{rule-based methods}, \textit{supervised methods} and \textit{semi-supervised methods}. 
For both semi-supervised and supervised methods, 
we use hard-matched sentences as the training data for comparing under the same setting.
The experiments evaluate the ability to use rules for improving RE. 

\smallskip
\noindent
\textbf{Rule-based Baseline Methods.} We apply the following models on $\mathcal{S}_\text{matched} \cup \mathcal{S}_\text{unmatched}$: {(1)} \texttt{CBOW-GloVe} adopts continuous bag-of-words ~\cite{mikolov2013distributed} on GloVe embeddings~\cite{pennington2014glove} to represent a sentence or rule body, which labels a sentence using its most similar rule (in cosine distance). 
\quad{(2)} \texttt{BREDS} \cite{batista2015semi} is a rule-based bootstrapping method originally designed for corpus-level RE. Given some entity pairs as seeds, it alternates between extracting new rules and new entity pairs. BREDS performs soft matching between rules and sentences using average word embeddings. In our experiments, we apply the set of rules learned by BREDS to perform hard-matching over the test sentences at the prediction time. 
\quad(3) The Neural Rule Engine (\texttt{NRE} \cite{li2018generalize}) is an unsupervised method of soft-matching. It generalizes given rules by first performing unigram matching for each token in the rules using CNN, then accumulates the matching scores of all tokens along the parsed tree structure. When used for prediction, NRE performs soft-matching on the given rules.

\smallskip
\noindent
\textbf{Supervised Baseline Methods.} We apply the following methods on $\mathcal{S}_\text{matched}$: (1) \texttt{PCNN} \cite{zeng2015distant} represents each token using both word embeddings and positional embeddings. Convolution and piecewise max pooling layer are performed to produce the sentence embedding.
\quad(2) \texttt{LSTM+ATT}  adopts bi-directional LSTM and attention mechanism~\cite{bahdanau2014neural} to produce a sentence embedding, which is fed into a fully-connected layer and a softmax classifier to predict the relation.
\quad(3) \texttt{PA-LSTM} \cite{zhang2017position} extends the LSTM model by incorporating positional information into attention mechanism and achieved state-of-the-art performance on TACRED.
\quad(4) \texttt{Data Programming} \cite{ratner2016data, Hancock_2018} denoises the conflicting rules by learning their accuracy and correlation structures.
\quad(5) \texttt{LSTM+ATT $\mathbf{(\mathcal{S}_\text{matched} + \mathcal{P})}$} extends the LSTM+ATT model by also using rules as training data. It serves as the base model for all semi-supervised baselines.

\smallskip
\noindent
\textbf{Semi-Supervised Baseline Methods.} 
To make fair comparisons with \NERO, we also regard labeling rules as training data (same to $L_\text{rules}$) and use \texttt{LSTM+ATT $\mathbf{(\mathcal{S}_\text{matched} + \mathcal{P})}$} as the base model. We apply the following methods on $\mathcal{S}_\text{matched} \cup \mathcal{S}_\text{unmatched} \cup \mathcal{P}$: (1) \texttt{Pseudo-Labeling} \cite{lee2013pseudo} trains the network using labeled and unlabeled data simultaneously. Pseudo-Labels are created for unlabeled data by picking up the class with the maximum predicted probability and are used as if they are labeled data during training. 
\quad(2) \texttt{Self-Training} \cite{rosenberg2005semi} iteratively trains the model using the labeled dataset and expands the labeled set using the most confident predictions among the unlabeled set. This procedure stops when unlabeled data is exhausted. 
\quad(3) \texttt{Mean-Teacher} \cite{tarvainen2017mean} assumes that data points with small differences should have similar outputs. We perturb each unlabeled sentence using word dropout and regularize their outputs to be similar. 
\quad(4) \texttt{DualRE} \cite{lin2019dualre} jointly trains a relation prediction and a retrieval module which mutually enhance each other by selecting high-quality instances from unlabeled data.

\smallskip
\noindent
\textbf{Variants of \NERO.} \quad(1) \texttt{\NERO w/o $\mathbf{\mathcal{S}_\text{unmatched}}$} removes the loss on unmatched sentences and only use $\mathcal{S}_\text{matched}$ and $\mathcal{P}$ in model training. It only performs hard-matching on rules. \quad(2) \texttt{\NERO-\texttt{SRM}} has the same training objective as \NERO, but uses the soft rule matcher (\texttt{SRM}) to make predictions. Given a sentence, \NERO-\texttt{SRM} finds the most similar rule and returns the rule head as the prediction. It can be taken as a context-agnostic version of \NERO (Sec.~\ref{sec::infer}). 

\subsection{Experiment Settings}
\label{app:impl}
\noindent \textbf{Implementation.} We implement most baselines from the scratch using Tensorflow 1.10 \cite{tensorflow2015-whitepaper} except for those that have released their codes (like \texttt{PA-LSTM} and \texttt{DualRE}). We adapt the baselines to our setting. For supervised and semi-supervised baselines, we use the hard-matched sentences ($\mathcal{S}_\text{matched}$) as the ``labeled data'' and the unmatched sentences ($\mathcal{S}_\text{unmatched}$) as the ``unlabeled data''.

\begin{table}[!t]
\vspace{-0.1cm}
\hspace{-0.4cm}
\scalebox{0.65}{
    \begin{tabular}{p{3.4cm}cccccc}
         \toprule
         \textbf{Method / Dataset} & \multicolumn{3}{c}{\textbf{TACRED}} & \multicolumn{3}{c}{\textbf{SemEval}} \\
          & Precision & Recall & $F_1$ & Precision & Recall & $F_1$ \\
         \midrule
         Rules &  85.0&  11.4&  20.1& 81.2 & 17.2 & 28.5 \\
         BREDS \cite{batista2015semi} & 53.8& 20.3& 29.5& 62.0 & 24.5 & 35.1 \\
         CBOW-GloVe & 27.9& 45.7& 34.6& 44.0& 52.8& 48.0 \\ 
         \midrule
         NRE~\cite{li2018generalize} & 65.2& 17.2& 27.2& 78.6 & 18.5 &30.0\\
         PCNN \cite{zeng2015distant} & 44.5 $\pm$ 0.4& 24.1 $\pm $ 2.8& 31.1 $\pm$ 2.6& 59.1 $\pm$ 1.4 & 43.0 $\pm$ 0.7 & 49.8 $\pm$ 0.5\\
         LSTM+ATT & 38.1 $\pm$ 2.7& 39.6 $\pm$ 2.7& 38.8 $\pm$ 2.4 & 64.5 $\pm$ 2.8& 53.3 $\pm$ 2.8& 58.2 $\pm$ 0.8 \\
         PA-LSTM \cite{zhang2017position} &  39.8 $\pm$ 2.5& 40.2 $\pm$ 2.0& 39.0 $\pm$ 0.6& 64.0 $\pm$ 3.6& 54.2 $\pm$ 2.5& 58.5 $\pm$ 0.6\\
         Data Programming \cite{ratner2016data}& 39.2 $\pm$ 1.3 & 40.1 $\pm$ 2.0& 39.7 $\pm$ 0.9& 61.8 $\pm$ 2.1& 54.8 $\pm$ 1.1& 58.1 $\pm$ 0.7\\
        \midrule
         LSTM+ATT ($\mathcal{S}_\text{matched}+\mathcal{P}$)& 39.2 $\pm$ 1.7& 45.5 $\pm$ 1.7 & 42.1 $\pm$ 0.9 & 63.4 $\pm$ 2.1& 55.0 $\pm$ 0.3& 58.8 $\pm$ 0.9\\ 
         Pseudo-Labeling \cite{lee2013pseudo}& 34.5 $\pm$ 4.1 & 37.4 $\pm$ 5.1 & 35.3 $\pm$ 0.8 &59.4 $\pm$ 3.3& 55.8 $\pm$ 2.1& 57.4 $\pm$ 1.3\\
         Self-Training \cite{rosenberg2005semi} &	37.8 $\pm$ 3.5&	41.1 $\pm$ 3.1&	39.2 $\pm$ 2.1& 62.3 $\pm$ 2.0& 53.0 $\pm$ 2.7& 57.1 $\pm$ 1.0 \\
         Mean-Teacher \cite{tarvainen2017mean} & 46.0 $\pm$ 2.7& 41.6 $\pm $ 2.2& 43.6 $\pm$ 1.3 & 62.3 $\pm$ 1.5 & 54.5 $\pm$ 1.2 & 57.9 $\pm$ 0.5 \\
         DualRE \cite{lin2019dualre} & 40.2 $\pm$ 1.5& 42.8 $\pm$ 2.0& 41.7 $\pm$ 0.5 & 63.7 $\pm$ 2.8 & 54.6 $\pm$ 2.1 & 58.6 $\pm$ 0.8 \\ \midrule
         NERO w/o $\mathcal{S}_\text{unmatched}$ &	41.9 $\pm$ 1.8&	44.3 $\pm$ 3.8&	42.9 $\pm$ 1.4& 61.4 $\pm$ 2.4& 56.2 $\pm$ 1.9& 58.6 $\pm$ 0.6 \\
         NERO-\texttt{SRM} &	45.6 $\pm$ 2.2&	45.2 $\pm$ 1.2&	45.3 $\pm$ 1.0& 54.8 $\pm$ 1.6& 55.2 $\pm$ 2.0& 54.9 $\pm$ 0.6 \\
         NERO & 54.0 $\pm$ 1.8& 48.9 $\pm$ 2.2& \textbf{51.3 $\pm$ 0.6}& 66.0 $\pm$ 1.5& 55.8 $\pm$ 0.9& \textbf{60.5 $\pm$ 0.7}\\
         \bottomrule
    \end{tabular}}
    \vspace{+0.2cm}
    \caption{\textbf{Performance comparison (in~\%) of relation extraction on the TACRED and SemEval datasets}. We report the mean and standard deviation of the evaluation metrics by conducting 5 runs of training and testing using different random seeds. We use LSTM+ATT ($\mathbf{\mathcal{S}_\text{matched} + \mathcal{P}}$) as the base model for all semi-supervised baselines and our models.}
    \vspace{-0.4cm}
    \label{tab::weakly_supervised}
\end{table}

\noindent \textbf{Training details.} We use pre-trained Glove embeddings \cite{pennington2014glove} to initialize the word embeddings and fine-tune them during training. For \NERO, We set the batch size to 50 for hard-matched sentences, and 100 for unmatched sentences. For other baselines, we set the batch size to 50. To exclude the possibility that \NERO takes advantage of a larger batch size, we also tried a batch size of 150 for other baselines but it showed no difference in performance. To reduce over-fitting, we adopt the entity masking technique \cite{zhang2017position} and replace the subject and object with \textsc{subj/obj--ner} and regard them as normal tokens. We use a two-layer bi-directional LSTM as the encoder. The hidden dimension is 100 for the LSTM, and 200 for the attention layers. We set $\beta$, $\gamma$, $\tau$ to 0.05, 0.5, and 1.0 respectively. We set $\alpha$ to 1.0 for TACRED, and 0.1 for SemEval. The temperature $\theta$ in instance weighting is set to 10. We apply dropout \cite{srivastava2014dropout} after the LSTM with a rate of 0.5. All models are optimized using AdaGrad \cite{duchi2011adaptive} with initial learning rate 0.5 and decay rate 0.95. For models that require the prediction of \textsc{None} (including \NERO and all supervised / semi-supervised baselines), we select the threshold (see \ref{sec::infer}) from range [0.0, 1.0] based on the dev set.

\subsection{Performance Comparison}
\noindent \textbf{Rule-based Models}. We first compare the models that solely use rules, as shown in Table~\ref{tab::weakly_supervised}.
Hard-matching with rules achieves a high precision (85\% on TACRED) but suffers from the low-recall problem due to its failure in matching instances with similar meanings (only 11.4\% on TACRED).
Bootstrapping methods (such as \texttt{BREDS}) and unsupervised soft-matching methods (such as \texttt{NRE}) manage to cover more instances (6\% gain in recall on TACRED dataset), but they fail to capture contextual information.

\smallskip
\noindent 
\textbf{Models Trained on $\mathcal{S}_\text{matched}$}. Neural networks are capable of fitting the hard-matched data, but they suffer from over-fitting due to the small data size. Adding the pattern encoder loss ($L_\text{rules}$) and the soft matching loss ($L_\text{clus}$) boosts the performance by a huge margin, as indicated by LSTM+ATT ($\mathcal{S}_\text{matched}+\mathcal{P}$) and \NERO w/o $\mathcal{S}_\text{unmatched}$). However, their performance is still far from satisfactory, since they cannot utilize large amounts of unmatched sentences. Data programming does not bring any improvement because our rule mining and labeling method rarely introduces conflicted labeling rules. Hence, data programming with our rules is the same as LSTM+ATT.

\begin{figure}
    \centering
    \includegraphics[width=0.85\linewidth]{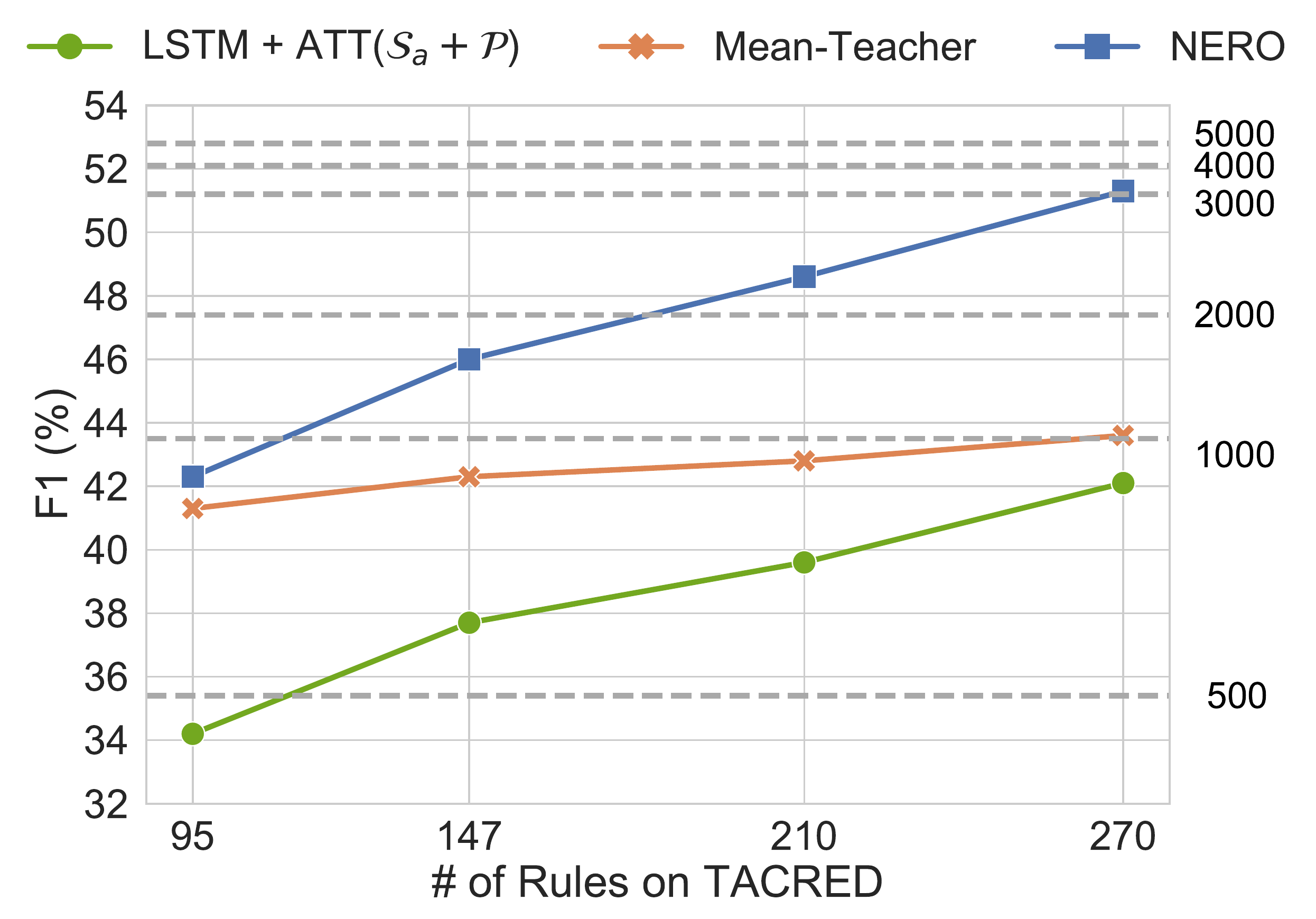}
    \vspace{-0.2cm}
    \caption{Performance w.r.t. different number of rules and human-annotated labels on TACRED. We show different models' F1 scores and number of rules or labels used for training the corresponding model.}
    \label{fig::rules}
    \vspace{-0.2cm}
\end{figure}

\smallskip 
\noindent 
\textbf{Models Trained on $\mathcal{S}_\text{matched} \cup \mathcal{S}_\text{unmatched} \cup \mathcal{P}$}. Table \ref{tab::weakly_supervised} shows the results for methods that incorporate unmatched data and rules.
For methods that actively create labels for unmatched data, namely pseudo-labeling, self-training and DualRE, their performance is even worse than the supervised counterparts (LSTM+ATT ($\mathcal{S}_\text{matched} + \mathcal{P}$)). It is because in a low-resource setting, the model performance is so low that the created labels are very noisy.
Mean-Teacher manages to improve model performance on TACRED, but only by a small margin.
Our proposed model, \NERO, is relatively indifferent to the noise due to the soft rule matcher, which learns directly from similar textual sentences. This results in a significant gain in precision (14\% on TACRED) and recall (4\% on TACRED) over the semi-supervised baselines.  Compared to TACRED, the improvement on SemEval is much smaller (only 1.7\% in F1). We hypothesize it is because the sentences in SemEval are quite short and contain very simple rule patterns. Thus, the soft-matched sentences can hardly provide additional information. We also try to predict with the soft rule matcher, but the performance is lower in F1 (45.3\% on TACRED and 54.9\% on SemEval) since it cannot capture contextual information.

\subsection{Performance Analysis}
\label{sec::performance_analysis}
\noindent \textbf{1. Performance on Various Amounts of Rules and Human Annotated Labels.} To show that our rules are more powerful than human-annotated labels, we show the performance of models trained with different number of rules (30\%, 50\%, 70\%, and 100\%) and the number of labels required to reach comparable performance using supervised baseline (LSTM+ATT). We random sample subsets of rules w.r.t. each relation to avoid extreme cases where some relations are not sampled at all. We conduct 5 runs of sampling and training and report the average performance. As shown in Figure \ref{fig::rules}, \NERO consistently outperforms other methods by a huge margin, and its performance increases with the number of rules (from 42.3\% to 51.3\%). Mean-Teacher also outperforms the base model especially when the number of rules is small, but its performance does not increase much (from 41.3\% to 43.6\%). It shows that Mean-Teacher generalizes well in low-resource setting but cannot fully utilize the knowledge in rules.
In terms of label efficiency, one rule in \NERO is equivalent to 10 human-annotated labels in model performance. Even in the base model LSTM+ATT ($\mathcal{S}_\text{matched} + \mathcal{P}$), one rule is still equivalent to about 4 labels. It demonstrates the superior generalization ability of rules over labels.

\begin{figure}
\vspace{-0.1cm}
    \centering
    \includegraphics[width=0.76\linewidth]{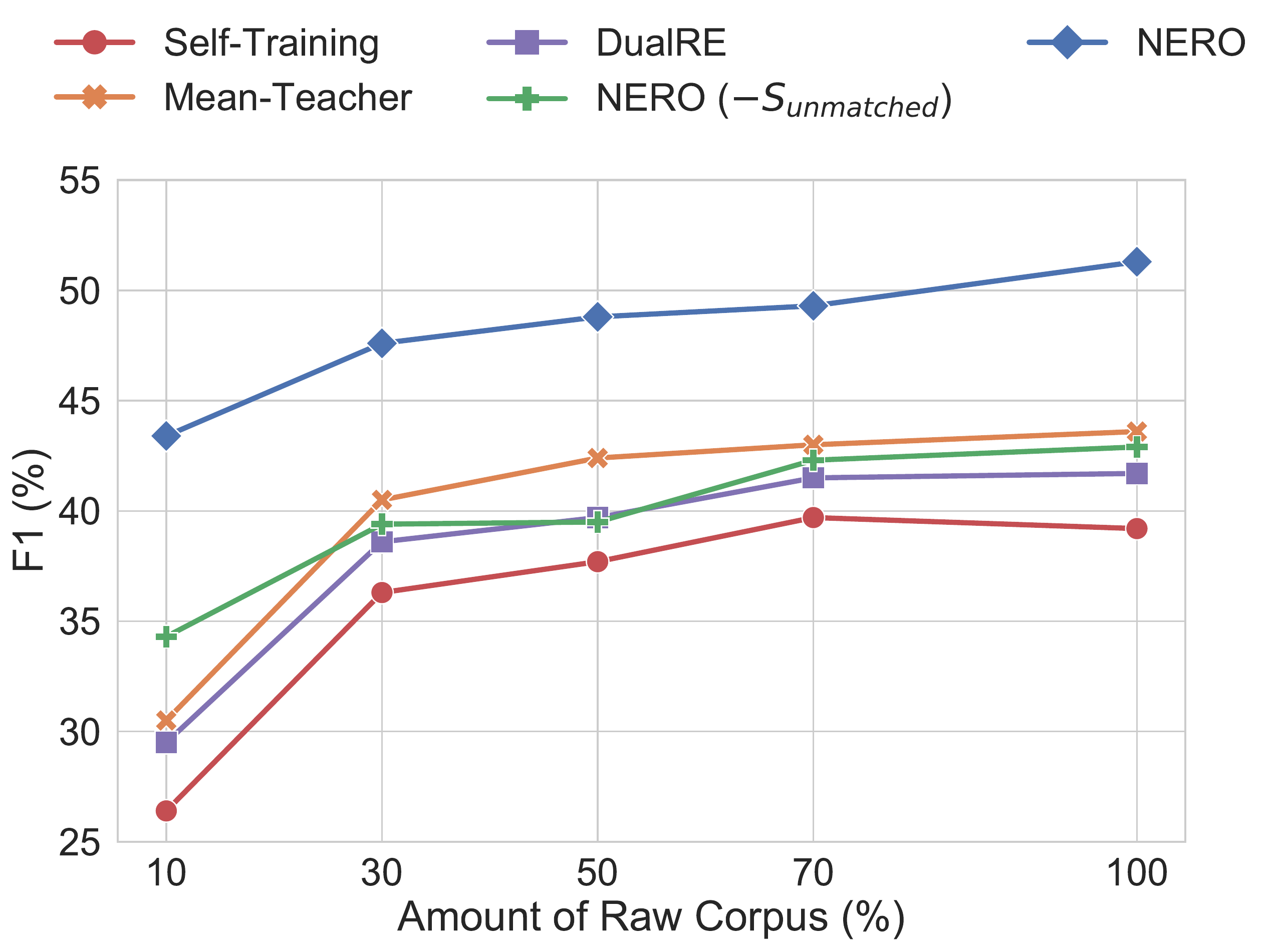}
    \vspace{-0.3cm}
    \caption{Performance of different semi-supervised models trained using various amounts of unlabeled sentences randomly sampled from the raw corpus on TACRED.}
    \label{fig::linechart}
    \vspace{-0.2cm}
\end{figure}

\smallskip
\noindent 
\textbf{2. Performance on Various Amounts of the Raw Corpus}. To test the robustness of our model, we plot the performance curve in Fig. \ref{fig::linechart} when different amounts of raw corpus are available on TACRED. Again, \NERO outperforms all methods, and its $F_1$ score positively correlates to the amount of available data (from 43.4\% to 51.3\%). We also observe similar phenomena in other methods (all based on LSTM+ATT ($\mathcal{S}_\text{matched} + \mathcal{P}$) model), which show that matching on rules provides additional knowledge to model training. Just memorizing these rules with a neural model leads to very bad performance. Also, we observe that with only 10\% of data, our final model (\NERO) already outperforms the best supervised baseline \NERO w/o $\mathcal{S}_\text{unmatched}$. This indicates that the soft matching module can utilize the raw corpus and rules in a more data-efficient way.

\smallskip
\noindent \textbf{3. Performance on Unseen Relations}. To show that \NERO has the capacity of predicting unseen relations, we evaluate \NERO on relations unseen in training. 
In the experiment, we randomly sample 5 relations as unseen relations and repeat the experiment 10 times. In each time, we use the same sets of unseen relations and the same random seed for all methods. We remove the sampled relations (both sentences and rules) in training while keeping only them in testing. We sample sentences of other relations as \textsc{None} relation in testing with the same ratio as the raw corpus (79.5\% for TACRED and 17.4\% for SemEval). Moreover, since the special tokens \textsc{subj/obj-ner} are out-of-vocabulary in both Glove and BERT, we only use the words between two entities in encoding and only use rules with the same entity types to the sentence in inference.
To make predictions by \NERO, we find the most similar rule using soft rule matcher.
Baseline models include exact rule matching, cosine similarity of average Glove embedding, and cosine similarity of pre-trained BERT embedding~\cite{devlin2018bert}.
As shown in Table \ref{tab::few_shot}, our soft rule matcher achieves similar or better performance compared to all baselines. It shows that instead of just memorizing the rules and the corresponding relation types, the soft rule matcher also learns knowledge about how to better perform rule matching and can even generalize to unseen relations.

\begin{table} [t]
\centering
\scalebox{0.87}{
\begin{tabular}{p{2.9cm}cccccc}
     \toprule
      & \multicolumn{3}{c}{TACRED} & \multicolumn{3}{c}{SemEval} \\
      Method & P & R & $F_1$ & P & R & $F_1$ \\
     \midrule
     Rule~(exact match) &100&6.1&10.8&83.2&17.7&28.2 \\
     CBOW-GloVe &52.4&86.3&64.7&40.3&45.5&34.7\\
     BERT-base (frozen)& 66.2& 76.8& \textbf{69.5}&37.8& 33.2& 35.3\\
     NERO &61.4 & 80.5& 68.9& 43.0& 54.1& \textbf{45.5} \\
     \bottomrule
\end{tabular}}
\vspace{0.1cm}
\caption{\textbf{Performance on predicting unseen relations.} \NERO applies the learned soft rule matcher on unseen relation rules to make predictions.}
\label{tab::few_shot}
\vspace{-0.4cm}
\end{table}

\subsection{Model Ablation Study}
\noindent
\textbf{1. Effect of Different Loss Functions}. To show the efficacy of our proposed losses ($L_\text{clus}$, $L_\text{unmatched}$ and $L_\text{rules}$), we conduct ablation study on loss functions by removing one loss function at a time. As shown in Table \ref{tab::ablation_loss}, all three losses contribute to the final performance, while $L_\text{unmatched}$ helps the most, which proves the effectiveness of soft-matching. When removing the contrastive loss $L_\text{clus}$, the F1 score drops from 51.3 to 46.4, which shows the effectiveness of a trainable \textsc{SRM}. The $L_\text{rules}$ loss also brings improvement, similar to including more training instances.

\begin{table}[h!]
\vspace{-0.1cm}
\centering
\scalebox{0.9}{
\begin{tabular}{p{1.6cm}ccc}
     \toprule
     Objective & Precision & Recall & $F_1$ \\
     \midrule
     $L$ (ours) & 54.0& 48.9& 51.3 \\
     $-L_\text{rules}$ & 50.0 & 47.7 & 49.0 \\
     $-L_\text{clus}$ & 50.9 & 43.0 & 46.4 \\
     $-L_\text{unmatched}$ & 41.9& 44.3& 42.9 \\
     \bottomrule
\end{tabular}}
\vspace{+0.1cm}
\caption{Ablation Study of Different Training Objectives on TACRED dataset. We remove each loss term one at a time.}
\label{tab::ablation_loss}
\vspace{-0.5cm}
\end{table}

\smallskip
\noindent \textbf{2. Effect of Different Soft-matching Models.} Besides word-level attention, we also try other soft matching mechanisms including CBOW-Glove, LSTM+ATT, and BERT-base (both frozen and fine-tuned). All methods first summarize the sentences / rules into vector representations, then calculate the matching scores by cosine similarity. We report their performance in Table \ref{tab::matcher}. We observe that our method achieves the best performance despite its simplicity. To our surprise, BERT-base performs much worse than our method. In the experiment we find that BERT tends to given high matching scores for all sentence-rule pairs so our framework cannot distinguish the false pseudo-labels from reliable ones.

\begin{table}[h!]
\vspace{-0.1cm}
\centering
\scalebox{0.88}{
\begin{tabular}{p{3.5cm}ccc}
     \toprule
     Objective & Precision & Recall & $F_1$\\
     \midrule
     CBOW-Glove & 49.4 & 43.5 & 46.2 \\
     LSTM+ATT & 56.2 & 46.0 & 50.6 \\
     BERT-base (frozen) &45.6& 47.6& 46.5\\
     BERT-base (fine-tuned) & 50.3& 45.8& 47.9 \\
     Word-level attention (ours) & 54.0& 48.9& 51.3\\
     \bottomrule
\end{tabular}}
\vspace{+0.1cm}
\caption{Ablation study of different soft-matching models for \NERO on the TACRED dataset.}
\label{tab::matcher}
\vspace{-0.4cm}
\end{table}

\begin{figure}[!b]
\vspace{-0.1cm}
    \centering
    \includegraphics[width=0.8\linewidth]{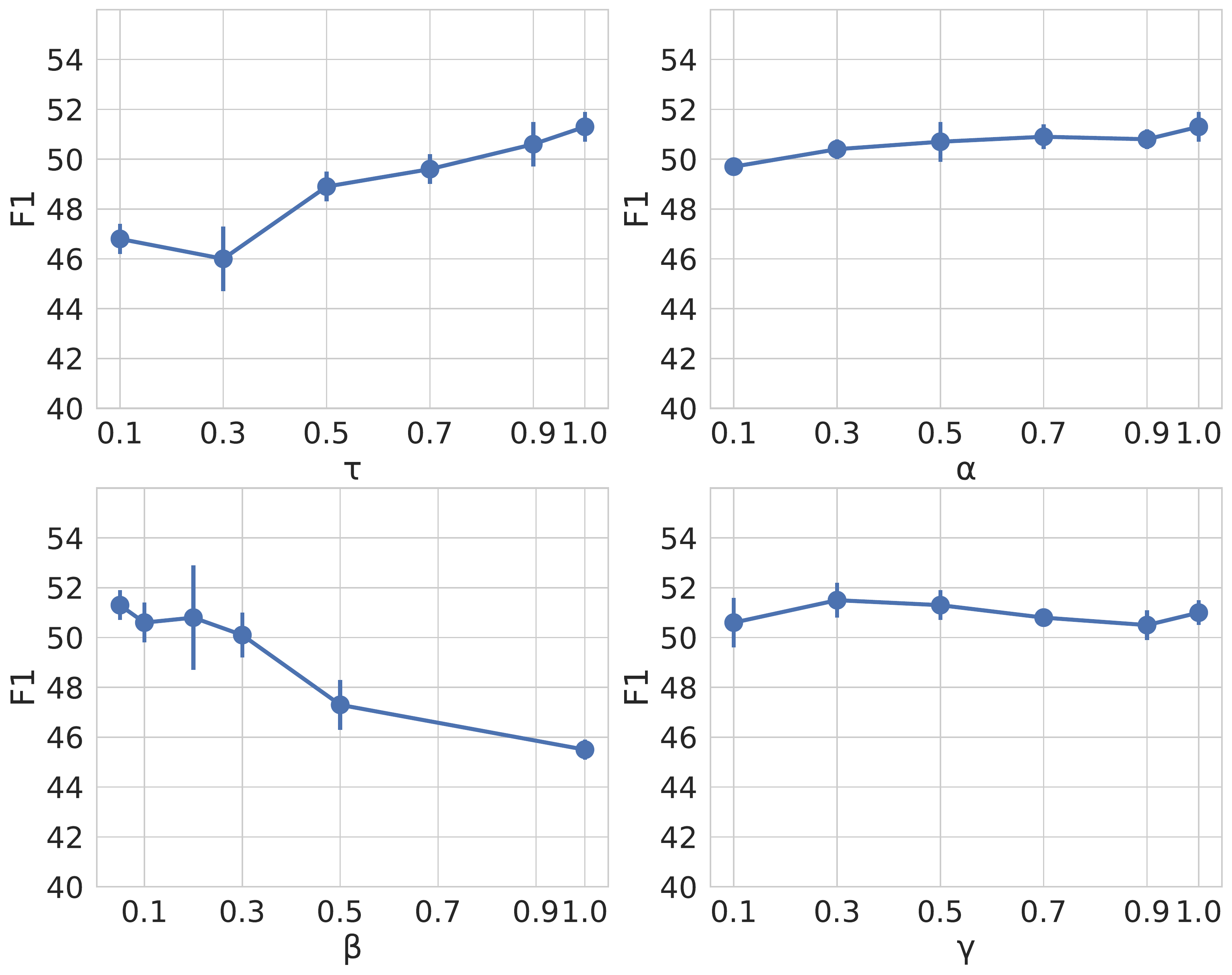}
    \vspace{-0.2cm}
    \caption{Sensitivity analysis of $\tau$, $\alpha$, $\beta$, and $\gamma$ on TACRED. We report the mean and standard deviation F1 score by conducting 5 runs of experiments using different random seeds.}
    \label{fig::sa}
    \vspace{-0.3cm}
\end{figure}

\smallskip
\noindent
\textbf{3. Sensitivity Analysis of Model Hyper-parameters.}
The most important hyper-parameters in \NERO are $\alpha$, $\beta$, $\gamma$, and $\tau$. We test the sensitivity of these parameters on TACRED. We adjust one hyper-parameter at a time and remain the other three unchanged. The results are shown in Figure \ref{fig::sa}.
We observe that for $\tau$ and $\beta$, there exists a wide interval where the F1 score remains stable. For $\alpha$ and $\gamma$, the F1 score hardly changes with the choice of their values. This is because that $L_\text{matched}$, $L_\text{clus}$, and $L_\text{rules}$ quickly drop to 0 and $L_\text{unmatched}$ dominates the joint loss in the training phase, and the impact of $\gamma$ is lessened by adaptive subgradient method (AdaGrad).

\subsection{Case Study}
\label{sec::case_study}

\begin{figure*}[!t]
\vspace{-0.2cm}
    \centering
    \includegraphics[width=0.92\linewidth]{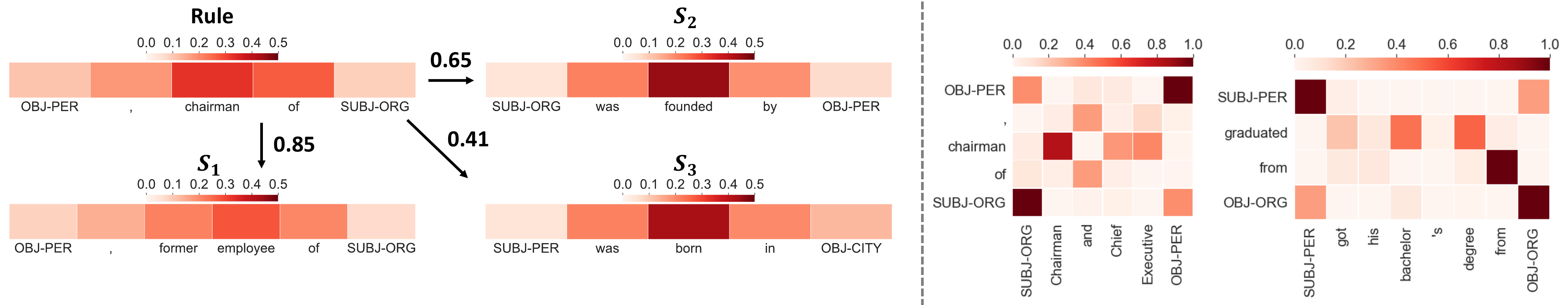}
    \vspace{-0.3cm}
    \caption{Output visualization of \texttt{SRM}. Left: attention weights of words and the soft matching scores between a rule and three sentences. Right: cosine similarity matrix between word embeddings learned with the contrastive loss.}
    \vspace{-0.2cm}
    \label{fig::heatmap}
\end{figure*}

\noindent
\textbf{Study on Label Efficiency.}
To test the label efficiency of rules and sentences in real scenarios, we ask 5 college students in computer science
to annotate frequent candidate rules (with frequency $\ge$ 3) and the unlabeled sentences, during a 40-minute period. For annotating sentences, they are required to provide labels from the pre-defined relation set $R \cup \textsc{None}$. For annotating candidate rules, they are required to filter out uninformative rules (e.g. ``\textsc{subj-person} and \textsc{obj-person}'') and assign labels to the remaining ones. For each candidate rule, they are further given an example of hard-matched sentence to help understand the meaning and context of the rule. Then we use each user's annotations to train the RE model. We use \textbf{LSTM+ATT} for labeled sentences and \textbf{\NERO} for rules.

\begin{figure}[h!]
    \centering
    \includegraphics[width=0.85\linewidth]{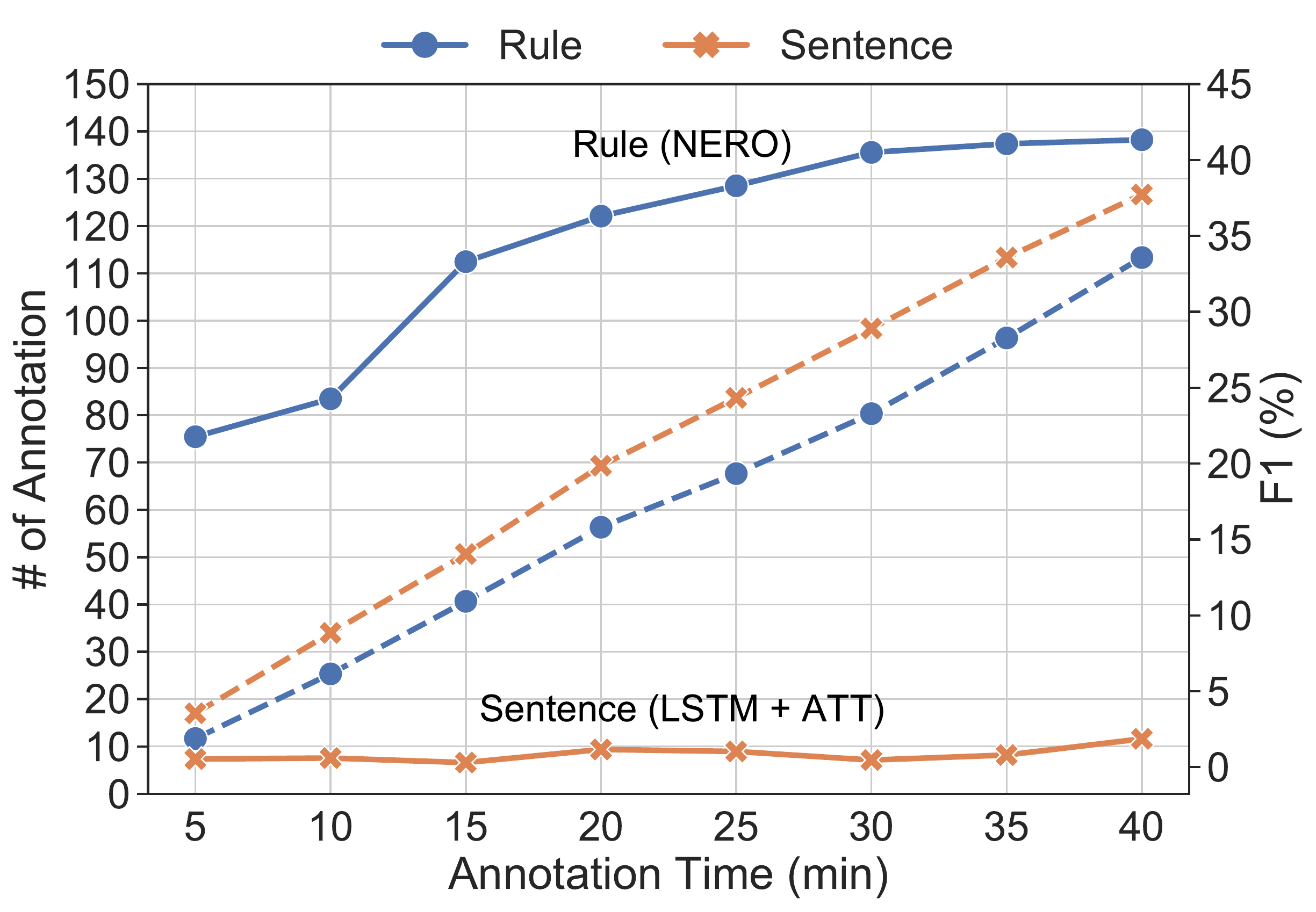}
    \vspace{-0.3cm}
    \caption{Study on label efficiency. Average number of rules / sentences labeled by annotators (dashed line) are shown on the x-axis over the left-hand side; and the performance of models trained with these corresponding labeled rules / sentences (solid line) are shown on the x-axis over the right-hand side. We use \NERO and LSTM+ATT as the base model for the labeling rules and sentences, respectively.}
    \label{fig::user_study}
    \vspace{-0.2cm}
\end{figure}

We report the average number of selected rules / labeled sentences and performance of trained RE model every 5 minutes in Figure \ref{fig::user_study}. We observe that getting rules is nearly as fast as getting labels but is much more efficient. With the same annotation time, \NERO can already learn a good RE classifier (with 41.3\% F1), while LSTM+ATT cannot learn anything from the labels. In just 40 minutes, \NERO can achieve comparable performance to 1000 labels (refer to Figure \ref{fig::rules}), which requires about 320 minutes to annotate.

\smallskip
\noindent
\textbf{Interpretability of Soft Rule Grounding.}
Fig.~\ref{fig::heatmap} gives an example of the similarity matrix, showing that our soft matching module learns meaningful representations of rules for soft matching.
The left part of the figure presents 1) which words are more important for building attention-pooled rule/sentence representations, and 2) the soft matching scores between the rule and three sentences.
We also show two concrete examples of attention matrix in the right part.
For each word-pair between a rule and a sentence, we visualize the cosine similarity matrix between  word representations learned with the contrastive loss.
While the word ``chairman'' in our rule matches perfectly with the same surface word in the sentence, it also has a high similarity score with ``Chief'' and ``Executive''.



\section{Related Work}
\noindent
\textbf{Relation Extraction}. 
Traditional corpus-level RE systems take a rule-based approach, in which rules are either handcrafted \cite{hearst1992automatic} or automatically learned from large corpora \cite{agichtein2000snowball, batista2015semi}. Although feature-based soft matching methods (e.g. TF-IDF, CBOW) are used in rule mining, the inference (i.e, labeling new sentences) is still done by hard-matching, which leads to high precision but low recall due to the low coverage of rules. While we use a trainable soft rule matcher, which is more powerful than prior feature-based methods. And we apply soft-matching to both training and inference phase in a consistent manner.
Recent RE models~\cite{zeng2015distant, zhang2017position} successfully apply deep neural networks with the help of large scale datasets such as SemEval \cite{hendrickx2009semeval} and TACRED \cite{zhang2017position}. However, their performance degrades heavily when labeled data is insufficient.
Some work \cite{Srivastava2017JointCL, Hancock_2018} proposes to train a RE classifier from natural language explanations, which are first transformed to labeling rules by a semantic parser then applied to raw corpora by hard-matching. But none of them utilizes the data that cannot be matched by rules, which is of massive amount.
This motivates us to propose a framework that can fully utilize the information from both rules and raw corpora.

\smallskip \noindent 
\textbf{Semi-Supervised Learning}.
Our work is relevant to semi-supervised learning, if we consider rule-annotated data as labeled among the corpus, e.g., self-training~\cite{rosenberg2005semi}, mean-teacher~\cite{tarvainen2017mean}, and semi-supervised VAE \cite{xu2017variational}.
In relation extraction, one line of work proposes to alleviate the low recall problem in rule-based approach using bootstrapping framework \cite{agichtein2000snowball} and revised pattern matching \cite{batista2015semi, gupta2018joint}. However, their settings are tailored for corpus-level RE. 
Another line of work applies self-training framework in supervised learning models. 
However, These models turn out to be ineffective in rule-labeled data due to potentially large difference in label distribution, and the generated psuedo-labels may be quite noisy.
\section{Conclusion}
In this paper, we proposed a novel framework, named \NERO, for label-efficient relation extraction.
We automatically extracted frequent patterns from large raw corpora and asked users to annotate them to form labeling rules.
To increase the coverage of these rules, we further proposed the soft-matching mechanism, where unlabeled sentences are annotated by their most semantically similar labeling rules and weighted in training the RE model.
Experiments on two public datasets proved the effectiveness of our framework.

\bibliographystyle{ACM-Reference-Format}
\bibliography{main}


\begin{thebibliography}{39}


\ifx \showCODEN    \undefined \def \showCODEN     #1{\unskip}     \fi
\ifx \showDOI      \undefined \def \showDOI       #1{#1}\fi
\ifx \showISBNx    \undefined \def \showISBNx     #1{\unskip}     \fi
\ifx \showISBNxiii \undefined \def \showISBNxiii  #1{\unskip}     \fi
\ifx \showISSN     \undefined \def \showISSN      #1{\unskip}     \fi
\ifx \showLCCN     \undefined \def \showLCCN      #1{\unskip}     \fi
\ifx \shownote     \undefined \def \shownote      #1{#1}          \fi
\ifx \showarticletitle \undefined \def \showarticletitle #1{#1}   \fi
\ifx \showURL      \undefined \def \showURL       {\relax}        \fi
\providecommand\bibfield[2]{#2}
\providecommand\bibinfo[2]{#2}
\providecommand\natexlab[1]{#1}
\providecommand\showeprint[2][]{arXiv:#2}

\bibitem[\protect\citeauthoryear{Abadi, Agarwal, Barham, Brevdo, Chen, Citro,
  Corrado, Davis, Dean, Devin, Ghemawat, Goodfellow, Harp, Irving, Isard, Jia,
  Jozefowicz, Kaiser, Kudlur, Levenberg, Man\'{e}, Monga, Moore, Murray, Olah,
  Schuster, Shlens, Steiner, Sutskever, Talwar, Tucker, Vanhoucke, Vasudevan,
  Vi\'{e}gas, Vinyals, Warden, Wattenberg, Wicke, Yu, and Zheng}{Abadi
  et~al\mbox{.}}{2015}]%
        {tensorflow2015-whitepaper}
\bibfield{author}{\bibinfo{person}{Mart\'{\i}n Abadi}, \bibinfo{person}{Ashish
  Agarwal}, \bibinfo{person}{Paul Barham}, \bibinfo{person}{Eugene Brevdo},
  \bibinfo{person}{Zhifeng Chen}, \bibinfo{person}{Craig Citro},
  \bibinfo{person}{Greg~S. Corrado}, \bibinfo{person}{Andy Davis},
  \bibinfo{person}{Jeffrey Dean}, \bibinfo{person}{Matthieu Devin},
  \bibinfo{person}{Sanjay Ghemawat}, \bibinfo{person}{Ian Goodfellow},
  \bibinfo{person}{Andrew Harp}, \bibinfo{person}{Geoffrey Irving},
  \bibinfo{person}{Michael Isard}, \bibinfo{person}{Yangqing Jia},
  \bibinfo{person}{Rafal Jozefowicz}, \bibinfo{person}{Lukasz Kaiser},
  \bibinfo{person}{Manjunath Kudlur}, \bibinfo{person}{Josh Levenberg},
  \bibinfo{person}{Dandelion Man\'{e}}, \bibinfo{person}{Rajat Monga},
  \bibinfo{person}{Sherry Moore}, \bibinfo{person}{Derek Murray},
  \bibinfo{person}{Chris Olah}, \bibinfo{person}{Mike Schuster},
  \bibinfo{person}{Jonathon Shlens}, \bibinfo{person}{Benoit Steiner},
  \bibinfo{person}{Ilya Sutskever}, \bibinfo{person}{Kunal Talwar},
  \bibinfo{person}{Paul Tucker}, \bibinfo{person}{Vincent Vanhoucke},
  \bibinfo{person}{Vijay Vasudevan}, \bibinfo{person}{Fernanda Vi\'{e}gas},
  \bibinfo{person}{Oriol Vinyals}, \bibinfo{person}{Pete Warden},
  \bibinfo{person}{Martin Wattenberg}, \bibinfo{person}{Martin Wicke},
  \bibinfo{person}{Yuan Yu}, {and} \bibinfo{person}{Xiaoqiang Zheng}.}
  \bibinfo{year}{2015}\natexlab{}.
\newblock \bibinfo{title}{{TensorFlow}: Large-Scale Machine Learning on
  Heterogeneous Systems}.
\newblock
\newblock
\urldef\tempurl%
\url{https://www.tensorflow.org/}
\showURL{%
\tempurl}
\newblock
\shownote{Software available from tensorflow.org.}


\bibitem[\protect\citeauthoryear{Agichtein and Gravano}{Agichtein and
  Gravano}{2000}]%
        {agichtein2000snowball}
\bibfield{author}{\bibinfo{person}{Eugene Agichtein} {and}
  \bibinfo{person}{Luis Gravano}.} \bibinfo{year}{2000}\natexlab{}.
\newblock \showarticletitle{Snowball: Extracting relations from large
  plain-text collections}. In \bibinfo{booktitle}{\emph{Proceedings of the
  fifth ACM conference on Digital libraries}}. ACM, \bibinfo{pages}{85--94}.
\newblock


\bibitem[\protect\citeauthoryear{Bahdanau, Cho, and Bengio}{Bahdanau
  et~al\mbox{.}}{2014}]%
        {bahdanau2014neural}
\bibfield{author}{\bibinfo{person}{Dzmitry Bahdanau},
  \bibinfo{person}{Kyunghyun Cho}, {and} \bibinfo{person}{Yoshua Bengio}.}
  \bibinfo{year}{2014}\natexlab{}.
\newblock \showarticletitle{Neural machine translation by jointly learning to
  align and translate}.
\newblock \bibinfo{journal}{\emph{arXiv preprint arXiv:1409.0473}}
  (\bibinfo{year}{2014}).
\newblock


\bibitem[\protect\citeauthoryear{Batista, Martins, and Silva}{Batista
  et~al\mbox{.}}{2015}]%
        {batista2015semi}
\bibfield{author}{\bibinfo{person}{David~S Batista}, \bibinfo{person}{Bruno
  Martins}, {and} \bibinfo{person}{M{\'a}rio~J Silva}.}
  \bibinfo{year}{2015}\natexlab{}.
\newblock \showarticletitle{Semi-supervised bootstrapping of relationship
  extractors with distributional semantics}. In
  \bibinfo{booktitle}{\emph{Proceedings of the 2015 Conference on Empirical
  Methods in Natural Language Processing}}. \bibinfo{pages}{499--504}.
\newblock


\bibitem[\protect\citeauthoryear{Devlin, Chang, Lee, and Toutanova}{Devlin
  et~al\mbox{.}}{2018}]%
        {devlin2018bert}
\bibfield{author}{\bibinfo{person}{Jacob Devlin}, \bibinfo{person}{Ming-Wei
  Chang}, \bibinfo{person}{Kenton Lee}, {and} \bibinfo{person}{Kristina
  Toutanova}.} \bibinfo{year}{2018}\natexlab{}.
\newblock \showarticletitle{BERT: Pre-training of Deep Bidirectional
  Transformers for Language Understanding}.
\newblock \bibinfo{journal}{\emph{arXiv preprint arXiv:1810.04805}}
  (\bibinfo{year}{2018}).
\newblock


\bibitem[\protect\citeauthoryear{Duchi, Hazan, and Singer}{Duchi
  et~al\mbox{.}}{2011}]%
        {duchi2011adaptive}
\bibfield{author}{\bibinfo{person}{John Duchi}, \bibinfo{person}{Elad Hazan},
  {and} \bibinfo{person}{Yoram Singer}.} \bibinfo{year}{2011}\natexlab{}.
\newblock \showarticletitle{Adaptive subgradient methods for online learning
  and stochastic optimization}.
\newblock \bibinfo{journal}{\emph{Journal of Machine Learning Research}}
  \bibinfo{volume}{12}, \bibinfo{number}{Jul} (\bibinfo{year}{2011}),
  \bibinfo{pages}{2121--2159}.
\newblock


\bibitem[\protect\citeauthoryear{Guo, Zhang, and Lu}{Guo et~al\mbox{.}}{2019}]%
        {guo2019attention}
\bibfield{author}{\bibinfo{person}{Zhijiang Guo}, \bibinfo{person}{Yan Zhang},
  {and} \bibinfo{person}{Wei Lu}.} \bibinfo{year}{2019}\natexlab{}.
\newblock \showarticletitle{Attention Guided Graph Convolutional Networks for
  Relation Extraction}.
\newblock \bibinfo{journal}{\emph{arXiv preprint arXiv:1906.07510}}
  (\bibinfo{year}{2019}).
\newblock


\bibitem[\protect\citeauthoryear{Gupta, Roth, and Sch{\"u}tze}{Gupta
  et~al\mbox{.}}{2018}]%
        {gupta2018joint}
\bibfield{author}{\bibinfo{person}{Pankaj Gupta}, \bibinfo{person}{Benjamin
  Roth}, {and} \bibinfo{person}{Hinrich Sch{\"u}tze}.}
  \bibinfo{year}{2018}\natexlab{}.
\newblock \showarticletitle{Joint bootstrapping machines for high confidence
  relation extraction}.
\newblock \bibinfo{journal}{\emph{arXiv preprint arXiv:1805.00254}}
  (\bibinfo{year}{2018}).
\newblock


\bibitem[\protect\citeauthoryear{Hancock, Varma, Wang, Bringmann, Liang, and
  Ré}{Hancock et~al\mbox{.}}{2018}]%
        {Hancock_2018}
\bibfield{author}{\bibinfo{person}{Braden Hancock}, \bibinfo{person}{Paroma
  Varma}, \bibinfo{person}{Stephanie Wang}, \bibinfo{person}{Martin Bringmann},
  \bibinfo{person}{Percy Liang}, {and} \bibinfo{person}{Christopher Ré}.}
  \bibinfo{year}{2018}\natexlab{}.
\newblock \showarticletitle{Training Classifiers with Natural Language
  Explanations}.
\newblock \bibinfo{journal}{\emph{Proceedings of the 56th Annual Meeting of the
  Association for Computational Linguistics (Volume 1: Long Papers)}}
  (\bibinfo{year}{2018}).
\newblock
\urldef\tempurl%
\url{https://doi.org/10.18653/v1/p18-1175}
\showDOI{\tempurl}


\bibitem[\protect\citeauthoryear{Hearst}{Hearst}{1992}]%
        {hearst1992automatic}
\bibfield{author}{\bibinfo{person}{Marti~A Hearst}.}
  \bibinfo{year}{1992}\natexlab{}.
\newblock \showarticletitle{Automatic acquisition of hyponyms from large text
  corpora}. In \bibinfo{booktitle}{\emph{Proceedings of the 14th conference on
  Computational linguistics-Volume 2}}. Association for Computational
  Linguistics, \bibinfo{pages}{539--545}.
\newblock


\bibitem[\protect\citeauthoryear{Hendrickx, Kim, Kozareva, Nakov,
  {\'O}~S{\'e}aghdha, Pad{\'o}, Pennacchiotti, Romano, and
  Szpakowicz}{Hendrickx et~al\mbox{.}}{2009}]%
        {hendrickx2009semeval}
\bibfield{author}{\bibinfo{person}{Iris Hendrickx}, \bibinfo{person}{Su~Nam
  Kim}, \bibinfo{person}{Zornitsa Kozareva}, \bibinfo{person}{Preslav Nakov},
  \bibinfo{person}{Diarmuid {\'O}~S{\'e}aghdha}, \bibinfo{person}{Sebastian
  Pad{\'o}}, \bibinfo{person}{Marco Pennacchiotti}, \bibinfo{person}{Lorenza
  Romano}, {and} \bibinfo{person}{Stan Szpakowicz}.}
  \bibinfo{year}{2009}\natexlab{}.
\newblock \showarticletitle{Semeval-2010 task 8: Multi-way classification of
  semantic relations between pairs of nominals}. In
  \bibinfo{booktitle}{\emph{Proceedings of the Workshop on Semantic
  Evaluations: Recent Achievements and Future Directions}}. Association for
  Computational Linguistics, \bibinfo{pages}{94--99}.
\newblock


\bibitem[\protect\citeauthoryear{Hochreiter and Schmidhuber}{Hochreiter and
  Schmidhuber}{1997}]%
        {hochreiter1997long}
\bibfield{author}{\bibinfo{person}{Sepp Hochreiter} {and}
  \bibinfo{person}{J{\"u}rgen Schmidhuber}.} \bibinfo{year}{1997}\natexlab{}.
\newblock \showarticletitle{Long short-term memory}.
\newblock \bibinfo{journal}{\emph{Neural computation}} \bibinfo{volume}{9},
  \bibinfo{number}{8} (\bibinfo{year}{1997}), \bibinfo{pages}{1735--1780}.
\newblock


\bibitem[\protect\citeauthoryear{Jiang and Zhai}{Jiang and Zhai}{2007}]%
        {jiang2007instance}
\bibfield{author}{\bibinfo{person}{Jing Jiang} {and}
  \bibinfo{person}{ChengXiang Zhai}.} \bibinfo{year}{2007}\natexlab{}.
\newblock \showarticletitle{Instance weighting for domain adaptation in NLP}.
  In \bibinfo{booktitle}{\emph{Proceedings of the 45th annual meeting of the
  association of computational linguistics}}. \bibinfo{pages}{264--271}.
\newblock


\bibitem[\protect\citeauthoryear{Jiang, Shang, Cassidy, Ren, Kaplan, Hanratty,
  and Han}{Jiang et~al\mbox{.}}{2017}]%
        {jiang2017metapad}
\bibfield{author}{\bibinfo{person}{Meng Jiang}, \bibinfo{person}{Jingbo Shang},
  \bibinfo{person}{Taylor Cassidy}, \bibinfo{person}{Xiang Ren},
  \bibinfo{person}{Lance~M Kaplan}, \bibinfo{person}{Timothy~P Hanratty}, {and}
  \bibinfo{person}{Jiawei Han}.} \bibinfo{year}{2017}\natexlab{}.
\newblock \showarticletitle{Metapad: Meta pattern discovery from massive text
  corpora}. In \bibinfo{booktitle}{\emph{Proceedings of the 23rd ACM SIGKDD
  International Conference on Knowledge Discovery and Data Mining}}. ACM,
  \bibinfo{pages}{877--886}.
\newblock


\bibitem[\protect\citeauthoryear{Jones, McCallum, Nigam, and Riloff}{Jones
  et~al\mbox{.}}{1999}]%
        {jones1999bootstrapping}
\bibfield{author}{\bibinfo{person}{Rosie Jones}, \bibinfo{person}{Andrew
  McCallum}, \bibinfo{person}{Kamal Nigam}, {and} \bibinfo{person}{Ellen
  Riloff}.} \bibinfo{year}{1999}\natexlab{}.
\newblock \showarticletitle{Bootstrapping for text learning tasks}. In
  \bibinfo{booktitle}{\emph{IJCAI-99 Workshop on Text Mining: Foundations,
  Techniques and Applications}}, Vol.~\bibinfo{volume}{1}.
\newblock


\bibitem[\protect\citeauthoryear{Lee}{Lee}{2013}]%
        {lee2013pseudo}
\bibfield{author}{\bibinfo{person}{Dong-Hyun Lee}.}
  \bibinfo{year}{2013}\natexlab{}.
\newblock \showarticletitle{Pseudo-label: The simple and efficient
  semi-supervised learning method for deep neural networks}.
\newblock


\bibitem[\protect\citeauthoryear{Li, Xu, and Lu}{Li et~al\mbox{.}}{2018}]%
        {li2018generalize}
\bibfield{author}{\bibinfo{person}{Shen Li}, \bibinfo{person}{Hengru Xu}, {and}
  \bibinfo{person}{Zhengdong Lu}.} \bibinfo{year}{2018}\natexlab{}.
\newblock \showarticletitle{Generalize Symbolic Knowledge With Neural Rule
  Engine}.
\newblock \bibinfo{journal}{\emph{arXiv preprint arXiv:1808.10326}}
  (\bibinfo{year}{2018}).
\newblock


\bibitem[\protect\citeauthoryear{Lin, Yan, Qu, and Ren}{Lin
  et~al\mbox{.}}{2019}]%
        {lin2019dualre}
\bibfield{author}{\bibinfo{person}{Hongtao Lin}, \bibinfo{person}{Jun Yan},
  \bibinfo{person}{Meng Qu}, {and} \bibinfo{person}{Xiang Ren}.}
  \bibinfo{year}{2019}\natexlab{}.
\newblock \showarticletitle{Learning Dual Retrieval Module for Semi-supervised
  Relation Extraction}. In \bibinfo{booktitle}{\emph{The Web Conference}}.
\newblock


\bibitem[\protect\citeauthoryear{Mikolov, Sutskever, Chen, Corrado, and
  Dean}{Mikolov et~al\mbox{.}}{2013}]%
        {mikolov2013distributed}
\bibfield{author}{\bibinfo{person}{Tomas Mikolov}, \bibinfo{person}{Ilya
  Sutskever}, \bibinfo{person}{Kai Chen}, \bibinfo{person}{Greg~S Corrado},
  {and} \bibinfo{person}{Jeff Dean}.} \bibinfo{year}{2013}\natexlab{}.
\newblock \showarticletitle{Distributed representations of words and phrases
  and their compositionality}. In \bibinfo{booktitle}{\emph{Advances in neural
  information processing systems}}. \bibinfo{pages}{3111--3119}.
\newblock


\bibitem[\protect\citeauthoryear{Mintz, Bills, Snow, and Jurafsky}{Mintz
  et~al\mbox{.}}{2009}]%
        {mintz2009distant}
\bibfield{author}{\bibinfo{person}{Mike Mintz}, \bibinfo{person}{Steven Bills},
  \bibinfo{person}{Rion Snow}, {and} \bibinfo{person}{Dan Jurafsky}.}
  \bibinfo{year}{2009}\natexlab{}.
\newblock \showarticletitle{Distant supervision for relation extraction without
  labeled data}. In \bibinfo{booktitle}{\emph{Proceedings of the Joint
  Conference of the 47th Annual Meeting of the ACL and the 4th International
  Joint Conference on Natural Language Processing of the AFNLP: Volume 2-Volume
  2}}. Association for Computational Linguistics, \bibinfo{pages}{1003--1011}.
\newblock


\bibitem[\protect\citeauthoryear{Nakashole, Weikum, and Suchanek}{Nakashole
  et~al\mbox{.}}{2012}]%
        {nakashole2012patty}
\bibfield{author}{\bibinfo{person}{Ndapandula Nakashole},
  \bibinfo{person}{Gerhard Weikum}, {and} \bibinfo{person}{Fabian Suchanek}.}
  \bibinfo{year}{2012}\natexlab{}.
\newblock \showarticletitle{PATTY: a taxonomy of relational patterns with
  semantic types}. In \bibinfo{booktitle}{\emph{Proceedings of the 2012 Joint
  Conference on Empirical Methods in Natural Language Processing and
  Computational Natural Language Learning}}. Association for Computational
  Linguistics, \bibinfo{pages}{1135--1145}.
\newblock


\bibitem[\protect\citeauthoryear{Neculoiu, Versteegh, and Rotaru}{Neculoiu
  et~al\mbox{.}}{2016}]%
        {Neculoiu2016LearningTS}
\bibfield{author}{\bibinfo{person}{Paul Neculoiu}, \bibinfo{person}{Maarten
  Versteegh}, {and} \bibinfo{person}{Mihai Rotaru}.}
  \bibinfo{year}{2016}\natexlab{}.
\newblock \showarticletitle{Learning Text Similarity with Siamese Recurrent
  Networks}. In \bibinfo{booktitle}{\emph{Rep4NLP@ACL}}.
\newblock


\bibitem[\protect\citeauthoryear{Pennington, Socher, and Manning}{Pennington
  et~al\mbox{.}}{2014}]%
        {pennington2014glove}
\bibfield{author}{\bibinfo{person}{Jeffrey Pennington},
  \bibinfo{person}{Richard Socher}, {and} \bibinfo{person}{Christopher
  Manning}.} \bibinfo{year}{2014}\natexlab{}.
\newblock \showarticletitle{Glove: Global vectors for word representation}. In
  \bibinfo{booktitle}{\emph{Proceedings of the 2014 conference on empirical
  methods in natural language processing (EMNLP)}}.
  \bibinfo{pages}{1532--1543}.
\newblock


\bibitem[\protect\citeauthoryear{Qu, Ren, Zhang, and Han}{Qu
  et~al\mbox{.}}{2018}]%
        {qu2018weakly}
\bibfield{author}{\bibinfo{person}{Meng Qu}, \bibinfo{person}{Xiang Ren},
  \bibinfo{person}{Yu Zhang}, {and} \bibinfo{person}{Jiawei Han}.}
  \bibinfo{year}{2018}\natexlab{}.
\newblock \showarticletitle{Weakly-supervised Relation Extraction by
  Pattern-enhanced Embedding Learning}. In
  \bibinfo{booktitle}{\emph{Proceedings of the 2018 World Wide Web Conference
  on World Wide Web}}. International World Wide Web Conferences Steering
  Committee, \bibinfo{pages}{1257--1266}.
\newblock


\bibitem[\protect\citeauthoryear{Ratner, De~Sa, Wu, Selsam, and R{\'e}}{Ratner
  et~al\mbox{.}}{2016}]%
        {ratner2016data}
\bibfield{author}{\bibinfo{person}{Alexander~J Ratner},
  \bibinfo{person}{Christopher~M De~Sa}, \bibinfo{person}{Sen Wu},
  \bibinfo{person}{Daniel Selsam}, {and} \bibinfo{person}{Christopher R{\'e}}.}
  \bibinfo{year}{2016}\natexlab{}.
\newblock \showarticletitle{Data programming: Creating large training sets,
  quickly}. In \bibinfo{booktitle}{\emph{Advances in neural information
  processing systems}}. \bibinfo{pages}{3567--3575}.
\newblock


\bibitem[\protect\citeauthoryear{Rosenberg, Hebert, and Schneiderman}{Rosenberg
  et~al\mbox{.}}{2005}]%
        {rosenberg2005semi}
\bibfield{author}{\bibinfo{person}{Chuck Rosenberg}, \bibinfo{person}{Martial
  Hebert}, {and} \bibinfo{person}{Henry Schneiderman}.}
  \bibinfo{year}{2005}\natexlab{}.
\newblock \showarticletitle{Semi-supervised self-training of object detection
  models}.
\newblock  (\bibinfo{year}{2005}).
\newblock


\bibitem[\protect\citeauthoryear{Roth, Barth, Wiegand, Singh, and Klakow}{Roth
  et~al\mbox{.}}{2014}]%
        {roth2014effective}
\bibfield{author}{\bibinfo{person}{Benjamin Roth}, \bibinfo{person}{Tassilo
  Barth}, \bibinfo{person}{Michael Wiegand}, \bibinfo{person}{Mittul Singh},
  {and} \bibinfo{person}{Dietrich Klakow}.} \bibinfo{year}{2014}\natexlab{}.
\newblock \showarticletitle{Effective slot filling based on shallow distant
  supervision methods}.
\newblock \bibinfo{journal}{\emph{arXiv preprint arXiv:1401.1158}}
  (\bibinfo{year}{2014}).
\newblock


\bibitem[\protect\citeauthoryear{Srivastava, Hinton, Krizhevsky, Sutskever, and
  Salakhutdinov}{Srivastava et~al\mbox{.}}{2014}]%
        {srivastava2014dropout}
\bibfield{author}{\bibinfo{person}{Nitish Srivastava},
  \bibinfo{person}{Geoffrey Hinton}, \bibinfo{person}{Alex Krizhevsky},
  \bibinfo{person}{Ilya Sutskever}, {and} \bibinfo{person}{Ruslan
  Salakhutdinov}.} \bibinfo{year}{2014}\natexlab{}.
\newblock \showarticletitle{Dropout: a simple way to prevent neural networks
  from overfitting}.
\newblock \bibinfo{journal}{\emph{The Journal of Machine Learning Research}}
  \bibinfo{volume}{15}, \bibinfo{number}{1} (\bibinfo{year}{2014}),
  \bibinfo{pages}{1929--1958}.
\newblock


\bibitem[\protect\citeauthoryear{Srivastava, Labutov, and Mitchell}{Srivastava
  et~al\mbox{.}}{2017}]%
        {Srivastava2017JointCL}
\bibfield{author}{\bibinfo{person}{Shashank Srivastava}, \bibinfo{person}{Igor
  Labutov}, {and} \bibinfo{person}{Tom~M. Mitchell}.}
  \bibinfo{year}{2017}\natexlab{}.
\newblock \showarticletitle{Joint Concept Learning and Semantic Parsing from
  Natural Language Explanations}. In \bibinfo{booktitle}{\emph{EMNLP}}.
\newblock


\bibitem[\protect\citeauthoryear{Surdeanu, Tibshirani, Nallapati, and
  Manning}{Surdeanu et~al\mbox{.}}{2012}]%
        {surdeanu2012multi}
\bibfield{author}{\bibinfo{person}{Mihai Surdeanu}, \bibinfo{person}{Julie
  Tibshirani}, \bibinfo{person}{Ramesh Nallapati}, {and}
  \bibinfo{person}{Christopher~D Manning}.} \bibinfo{year}{2012}\natexlab{}.
\newblock \showarticletitle{Multi-instance multi-label learning for relation
  extraction}. In \bibinfo{booktitle}{\emph{Proceedings of the 2012 joint
  conference on empirical methods in natural language processing and
  computational natural language learning}}. Association for Computational
  Linguistics, \bibinfo{pages}{455--465}.
\newblock


\bibitem[\protect\citeauthoryear{Tarvainen and Valpola}{Tarvainen and
  Valpola}{2017}]%
        {tarvainen2017mean}
\bibfield{author}{\bibinfo{person}{Antti Tarvainen} {and}
  \bibinfo{person}{Harri Valpola}.} \bibinfo{year}{2017}\natexlab{}.
\newblock \showarticletitle{Mean teachers are better role models:
  Weight-averaged consistency targets improve semi-supervised deep learning
  results}. In \bibinfo{booktitle}{\emph{Advances in neural information
  processing systems}}. \bibinfo{pages}{1195--1204}.
\newblock


\bibitem[\protect\citeauthoryear{Wang, Cao, de~Melo, and Liu}{Wang
  et~al\mbox{.}}{2016}]%
        {wang-etal-2016-relation}
\bibfield{author}{\bibinfo{person}{Linlin Wang}, \bibinfo{person}{Zhu Cao},
  \bibinfo{person}{Gerard de Melo}, {and} \bibinfo{person}{Zhiyuan Liu}.}
  \bibinfo{year}{2016}\natexlab{}.
\newblock \showarticletitle{Relation Classification via Multi-Level Attention
  {CNN}s}. In \bibinfo{booktitle}{\emph{Proceedings of the 54th Annual Meeting
  of the Association for Computational Linguistics (Volume 1: Long Papers)}}.
  \bibinfo{publisher}{Association for Computational Linguistics},
  \bibinfo{address}{Berlin, Germany}, \bibinfo{pages}{1298--1307}.
\newblock
\urldef\tempurl%
\url{https://doi.org/10.18653/v1/P16-1123}
\showDOI{\tempurl}


\bibitem[\protect\citeauthoryear{Willett}{Willett}{2006}]%
        {willett2006porter}
\bibfield{author}{\bibinfo{person}{Peter Willett}.}
  \bibinfo{year}{2006}\natexlab{}.
\newblock \showarticletitle{The Porter stemming algorithm: then and now}.
\newblock \bibinfo{journal}{\emph{Program}} \bibinfo{volume}{40},
  \bibinfo{number}{3} (\bibinfo{year}{2006}), \bibinfo{pages}{219--223}.
\newblock


\bibitem[\protect\citeauthoryear{Xu, Sun, Deng, and Tan}{Xu
  et~al\mbox{.}}{2017}]%
        {xu2017variational}
\bibfield{author}{\bibinfo{person}{Weidi Xu}, \bibinfo{person}{Haoze Sun},
  \bibinfo{person}{Chao Deng}, {and} \bibinfo{person}{Ying Tan}.}
  \bibinfo{year}{2017}\natexlab{}.
\newblock \showarticletitle{Variational autoencoder for semi-supervised text
  classification}. In \bibinfo{booktitle}{\emph{Thirty-First AAAI Conference on
  Artificial Intelligence}}.
\newblock


\bibitem[\protect\citeauthoryear{Yu and Dredze}{Yu and Dredze}{2015}]%
        {yu2015learning}
\bibfield{author}{\bibinfo{person}{Mo Yu} {and} \bibinfo{person}{Mark Dredze}.}
  \bibinfo{year}{2015}\natexlab{}.
\newblock \showarticletitle{Learning composition models for phrase embeddings}.
\newblock \bibinfo{journal}{\emph{Transactions of the Association for
  Computational Linguistics}}  \bibinfo{volume}{3} (\bibinfo{year}{2015}),
  \bibinfo{pages}{227--242}.
\newblock


\bibitem[\protect\citeauthoryear{Zeng, Liu, Chen, and Zhao}{Zeng
  et~al\mbox{.}}{2015}]%
        {zeng2015distant}
\bibfield{author}{\bibinfo{person}{Daojian Zeng}, \bibinfo{person}{Kang Liu},
  \bibinfo{person}{Yubo Chen}, {and} \bibinfo{person}{Jun Zhao}.}
  \bibinfo{year}{2015}\natexlab{}.
\newblock \showarticletitle{Distant supervision for relation extraction via
  piecewise convolutional neural networks}. In
  \bibinfo{booktitle}{\emph{Proceedings of the 2015 Conference on Empirical
  Methods in Natural Language Processing}}. \bibinfo{pages}{1753--1762}.
\newblock


\bibitem[\protect\citeauthoryear{Zhang, Qi, and Manning}{Zhang
  et~al\mbox{.}}{2018}]%
        {zhang2018graph}
\bibfield{author}{\bibinfo{person}{Yuhao Zhang}, \bibinfo{person}{Peng Qi},
  {and} \bibinfo{person}{Christopher~D Manning}.}
  \bibinfo{year}{2018}\natexlab{}.
\newblock \showarticletitle{Graph convolution over pruned dependency trees
  improves relation extraction}.
\newblock \bibinfo{journal}{\emph{arXiv preprint arXiv:1809.10185}}
  (\bibinfo{year}{2018}).
\newblock


\bibitem[\protect\citeauthoryear{Zhang, Zhong, Chen, Angeli, and Manning}{Zhang
  et~al\mbox{.}}{2017}]%
        {zhang2017position}
\bibfield{author}{\bibinfo{person}{Yuhao Zhang}, \bibinfo{person}{Victor
  Zhong}, \bibinfo{person}{Danqi Chen}, \bibinfo{person}{Gabor Angeli}, {and}
  \bibinfo{person}{Christopher~D Manning}.} \bibinfo{year}{2017}\natexlab{}.
\newblock \showarticletitle{Position-aware attention and supervised data
  improve slot filling}. In \bibinfo{booktitle}{\emph{Proceedings of the 2017
  Conference on Empirical Methods in Natural Language Processing}}.
  \bibinfo{pages}{35--45}.
\newblock


\bibitem[\protect\citeauthoryear{Zhou, Shi, Tian, Qi, Li, Hao, and Xu}{Zhou
  et~al\mbox{.}}{2016}]%
        {zhou-etal-2016-attention}
\bibfield{author}{\bibinfo{person}{Peng Zhou}, \bibinfo{person}{Wei Shi},
  \bibinfo{person}{Jun Tian}, \bibinfo{person}{Zhenyu Qi},
  \bibinfo{person}{Bingchen Li}, \bibinfo{person}{Hongwei Hao}, {and}
  \bibinfo{person}{Bo Xu}.} \bibinfo{year}{2016}\natexlab{}.
\newblock \showarticletitle{Attention-Based Bidirectional Long Short-Term
  Memory Networks for Relation Classification}. In
  \bibinfo{booktitle}{\emph{Proceedings of the 54th Annual Meeting of the
  Association for Computational Linguistics (Volume 2: Short Papers)}}.
  \bibinfo{publisher}{Association for Computational Linguistics},
  \bibinfo{address}{Berlin, Germany}, \bibinfo{pages}{207--212}.
\newblock
\urldef\tempurl%
\url{https://doi.org/10.18653/v1/P16-2034}
\showDOI{\tempurl}


\end{thebibliography}

\end{document}